%% file: main.tex
\documentclass[conference]{IEEEtran}

\usepackage{cite}
\usepackage{mathtools}
\usepackage{amsmath,amssymb,amsfonts}
\usepackage{algorithm}
\usepackage{algpseudocode}
\usepackage{graphicx}
\usepackage{subcaption}
\usepackage{textcomp}
\usepackage{xcolor}
\usepackage[a4paper, total={184mm,239mm}]{geometry}
\def\BibTeX{{\rm B\kern-.05em{\sc i\kern-.025em b}\kern-.08em
    T\kern-.1667em\lower.7ex\hbox{E}\kern-.125emX}}

\usepackage[LGRgreek, italic]{mathastext}
\usepackage{multirow,tabularx}
\usepackage{makecell}
\usepackage[font=small,labelfont=bf]{caption}

\graphicspath{ {images/} }

\begin{document}

 \title{DNA-TEQ: An Adaptive Exponential Quantization of Tensors for DNN Inference}
\author{\IEEEauthorblockN{Bahareh Khabbazan, Marc Riera, Antonio González }
\IEEEauthorblockA{\textit{dept. of Computer Architecture } \\
\textit{Universitat Polit\`{e}cnica de Catalunya (UPC)}\\
Barcelona, Spain\\
\{bahareh.khabbazan, marc.riera.villanueva, antonio.gonzalez\}@upc.edu}
}


\maketitle

\begin{abstract}
Quantization is commonly used in Deep Neural Networks (DNNs) to reduce the storage and computational complexity by decreasing the arithmetical precision of activations and weights, a.k.a. tensors. Efficient hardware architectures employ linear quantization to enable the deployment of recent DNNs onto embedded systems and mobile devices. However, linear uniform quantization cannot usually reduce the numerical precision to less than 8 bits without sacrificing high performance in terms of model accuracy. The performance loss is due to the fact that tensors do not follow uniform distributions. In this paper, we show that a significant amount of tensors fit into an exponential distribution. Then, we propose DNA-TEQ to exponentially quantize DNN tensors with an adaptive scheme that achieves the best trade-off between numerical precision and accuracy loss. The experimental results show that DNA-TEQ provides a much lower quantization bit-width compared to previous proposals, resulting in an average compression ratio of 40\% over the linear INT8 baseline, with negligible accuracy loss and without retraining the DNNs. Besides, DNA-TEQ leads the way in performing dot-product operations in the exponential domain. On average for a set of widely used DNNs, DNA-TEQ provides $1.5x$ speedup and $2.5x$ energy savings over a baseline DNN accelerator based on 3D-stacked memory.
\end{abstract}

\begin{IEEEkeywords}
DNN, Quantization, Exponential, Transformer
\end{IEEEkeywords}

\input{sections/1-introduction}
\input{sections/2-related_research}

\input{sections/3-proposed_method}
\input{sections/4-evaluation}
\input{sections/5-conclusion}
\input{sections/6-acknowledgement}

\bibliographystyle{IEEEtranS}
\bibliography{references}

\end{document}

%% file: sections/1-introduction.tex
\section{Introduction}
Deep Neural Networks (DNNs) have achieved human performance levels in cognitive computing applications such as speech recognition or machine translation~\cite{google_human}. In order to achieve these levels of accuracy, DNNs have evolved and grown in complexity. From small Multilayer Perceptrons (MLPs) for simple tasks like recognizing written digits or characters, passing by large Convolutional Neural Networks (CNNs)~\cite{AlexNetV2, ResNet} for recognizing objects in images, to complex Recurrent Neural Networks (RNNs)~\cite{recurrent} and Transformers to reach the aforementioned human parity in speech recognition and machine translation. Nowadays, the machine learning community has shifted the focus to Transformer models~\cite{attention, transpim, language} due to their extreme efficiency in terms of both accuracy and performance for multiple Natural Language Processing (NLP) applications. However, MLPs and CNNs still provide state-of-the-art accuracy for applications such as acoustic scoring in speech recognition or self-driving cars respectively~\cite{understanding, deeptest}. The dramatic growth of DNNs size has made their deployment on mobile and embedded devices extremely challenging~\cite{survey}.

Over the years, DNNs have grown steadily in computational complexity and memory footprint~\cite{convolutional}. For example, the Switch Transformer~\cite{switch_transformer} from Google has 1.6 trillion parameters, which requires 890 billion floating-point multiply-and-add operations per forward-pass and terabytes of memory to store the weights. The development trend of most DNN models shows that the more parameters, the more powerful the models become in terms of accuracy~\cite{mokey}. However, this comes at the cost of more storage, data transfers, and computations. Hardware architectures must meet the memory storage and computational needs of modern DNNs. As these models are huge and the on-chip memory of current chips is limited, it is off-chip memory accesses that account for most of the energy consumption~\cite{eie}. Therefore, techniques to compress or reduce the amount of parameters are required to perform efficient inference of DNNs in any given hardware architecture~\cite{algorithm_architecture}.

Researchers have proposed several DNN optimizations~\cite{mriera_thesis, permdnn, deep} to ease the execution of real-time applications on mobile and embedded devices. Based on the observation that DNN models tend to be oversized, pruning removes redundant weights and/or activations to compress the DNN model size and reduce the amount of computations. However, pruning requires a costly re-training procedure of the DNNs to recover the full accuracy, as well as complex hardware to efficiently decode and run sparse models. On the other hand, linear uniform quantization is a popular technique to map a continuous set of values to a finite set. The main benefits of uniform quantization are the reduction in storage, due to a smaller representation of the tensor values, and the lower computational complexity, due to performing integer operations rather than floating point. Uniform quantization typically reduces the arithmetical precision of activations and weights to 8 bits with negligible accuracy loss for many networks, but lower bitwidths may result in high errors~\cite{quantizing}. Despite the advantages of pruning and uniform quantization, modern DNN models demand more aggressive optimizations to be suitable for current hardware architectures.

Recently, aggressive non-uniform quantization schemes have been proposed to further reduce the memory footprint by lowering the precision under 8 bits. For example, logarithmic quantization (LQ)~\cite{L2L, convolutional, deep} has been previously proposed to represent weights and/or activations with only a fixed base and an integer exponent. LQ exploits the non-uniform distributions of tensors to achieve smaller bitwidths compared to the uniform quantization. Most works employ a base-2 LQ to not only compress the model size but also eliminate bulky digital multipliers by using simple add and shift operations. However, base-2 LQ is still far from being the quantization that best represents the distributions of tensors for modern DNNs, introducing large amounts of error that cannot be compensated even after re-training. Consequently, there is a need to further investigate in quantization methods that obtain the best trade-off between accuracy and computational complexity.

In this paper, we propose a novel DNN Adaptive Tensor Exponential Quantization methodology, named DNA-TEQ, to quantize DNN tensors with an exponential representation by considering their distributions. DNA-TEQ includes an adaptive offline search algorithm to find the optimal parameters that provide the best trade-off between numerical precision and DNNs accuracy on a layer and DNN basis. DNA-TEQ also exploits the exponential quantization to simplify the dot-product operations by counting exponents. To demonstrate the cost-effectiveness of the exponential dot-product operations, we implement and evaluate DNA-TEQ both in software, using vector instructions (i.e. SIMD), and in hardware, on top of a baseline 3D-stacked DRAM-based DNN accelerator. Our experimental results show that specialized hardware is required to fully support the execution of different DNN models, providing higher performance gains and energy savings than the software counterpart.

To summarize, the goal of this work is to propose a quantization methodology that 1) does not require re-training, 2) quantizes weights and activations together, 3) generalizes for different DNNs, and 4) reduces hardware complexity improving performance and energy efficiency. The main contributions are:

\begin{itemize}
\item We analyze the distributions of activations and weights of different DNN models. We observe that most tensors follow a non-uniform distribution close to an exponential.

\item We propose DNA-TEQ, an adaptive quantization scheme that finds the optimal parameters for the exponential representation of tensors. DNA-TEQ minimizes the quantization error and achieves the best trade-off between numerical precision and model accuracy for several DNNs. On average, DNA-TEQ provides 40\% compression over INT8 with negligible accuracy loss. Model parameters can be represented by only 3 bits in some cases.

\item We implement DNA-TEQ by software using Intel AVX-512 SIMD instructions for a set of fully-connected (FC) layers with various sizes. DNA-TEQ achieves up to $5x$ speedup over a highly optimized INT8 implementation employing Vector Neural Network Instructions (VNNI). However, we observe that SIMD does not adapt well to the mix of different numerical precisions supported by DNA-TEQ, limiting the potential benefits in many DNNs.

\item We present a hardware accelerator based on 3D-stacked memory to take full advantage of DNA-TEQ. We evaluate it for a set of representative DNNs. DNA-TEQ improves performance by $1.5x$ and reduces energy consumption by $2.5x$ on average over a baseline INT8 accelerator.
\end{itemize}

The rest of the paper is organized as follows. Section~\ref{related_works} provides a summary of works related to DNN quantization techniques. Section~\ref{Proposed_method} describes our proposed quantization methodology including an analysis of the range and distribution of activations and weights for a set of popular DNNs. Sections~\ref{simd_implementation}-\ref{accelerator_architecture} present the software and hardware implementation to exploit the benefits of our quantization scheme. Section~\ref{evaluation} provides the evaluation methodology and experimental results. Finally, Section~\ref{conclusion} sums up the main conclusions.

%% file: sections/2-related_research.tex




\section{Related Work}\label{related_works}
Popular optimizations~\cite{mriera_thesis} for reducing the memory footprint of DNNs include clustering and pruning. Clustering uses methods such as K-means to compress the number of different weights to K centroids. Each weight is then substituted by an index that corresponds to the closest centroid. Clustering alone does not reduce the amount of computations nor its cost, but storage requirements. On the other hand, pruning~\cite{marc_prune} reduces the model size and the number of computations by removing unimportant parameters. The pruned model may loss accuracy but tends to regain it after retraining. Nevertheless, the pruned model becomes sparse, requiring specialized hardware to be efficiently executed. Pruning is orthogonal to DNN quantization methods such as DNA-TEQ.

As previously discussed, uniform quantization~\cite{mriera_thesis, redy} has been used for DNNs in the past to reduce the numerical precision (i.e. 8b-16b) and computational cost with small impact in accuracy~\cite{quantizing}. However, the latest advances in machine learning have sparked the research of new quantization schemes to further reduce the memory footprint and computational complexity of modern DNN models~\cite{bit_eff_qt}. Some works~\cite{datafree, rokh2022comprehensive} have attempted to use uniform quantization to reduce the numerical precision below 8 bits without paying attention to the tensor distributions, resulting in huge accuracy drops when performed on complex neural networks.

In order to further reduce the model size and computational cost of DNNs, recent studies~\cite{song, khoram-adaptive} proposed to set different bitwidths for weights and/or activations based on their sensitivity to quantization error. As a result, multiple mixed precision quantization schemes have been developed. These schemes perform an analysis of each tensor distribution to use it as a criterion to determine the optimal numerical precision for each DNN layer. Mixed precision quantization schemes have the potential to represent the tensors with lower bitwidths (i.e. $<8b$). Even so, most of these schemes still use uniform quantization once the precision of each layer has been determined, which may impact the accuracy of the model. Similarly, DNA-TEQ employs a mixed precision scheme for each layer but with a non-uniform quantization representation.

Several works~\cite{convolutional, deep, L2L, pre-trained, 2022log_asap} proposed non-uniform quantization schemes based on the base-2 logarithmic representation of weights and/or activations of CNNs to reduce the numerical precision (i.e. $<8b$) and HW complexity. The work in \cite{deep} only quantizes the weights to avoid the overhead of the dynamic quantization of activations, but the logarithmic quantization of both can lead to better schemes to reduce the overall computational cost. In particular, the authors of \cite{convolutional} exploit the base-2 quantization of activations and weights to perform operations in log-domain. However, the reduction in accuracy is non-negligible even in simple networks.

APoT~\cite{additive} is a non-uniform quantization scheme for the bell-shaped and long-tailed distribution of weights and activations in DNNs. APoT constrains all quantization levels as the sum of powers-of-two terms. They can reduce the numerical precision of ResNet to 4 bits but require retraining to recover the accuracy. On the other hand, Mokey~\cite{mokey} is a post-training compression method for Transformer models that does not require retraining. Mokey reduces the memory footprint by quantizing all values to 4-bit indexes into dictionaries of representative 16-bit fixed-point centroids. Mokey selects centroid values to also fit an exponential curve from a random normal distribution in an offline step. However, their method can not be adapted to any DNN model but Transformers. Moreover, they require to compute outliers in a costly post-processing phase.

In contrast to these works, DNA-TEQ reduces both storage and computational cost by quantizing activations and weights together, does not require retraining, and can be easily adapted to any DNN. The implementation details of DNA-TEQ are described in the following section.

%% file: sections/3-proposed_method.tex
\section{DNA-TEQ}\label{Proposed_method}
This section describes the proposed methodology for applying exponential quantization with pseudo-optimal parameters on a variety of DNNs. As discussed in previous sections, DNN quantization allows to reduce the numerical precision of activations and weights, which in turn lowers the memory footprint and the hardware complexity of the functional units. On the other hand, a proper quantization method should consider the distribution of activations and weights to reduce the impact in accuracy loss by minimizing the quantization error. The optimal quantization scheme must find the best trade-off between numerical precision and DNNs accuracy. In this work, we first observe and compare different distributions of activations and weights and show that an exponential quantization is the best fit. Then, we propose an offline search algorithm to find the optimal base of the exponential quantization along with other essential parameters. Finally, we take advantage of the exponential quantization to transform the traditional linear dot-product operations of DNNs into the exponential domain, which simplifies the hardware required to perform DNN inference.

\subsection{Tensor Distribution Analysis}\label{distribution_analysis}
Previous works have used uniform quantization to compress the DNN parameters. However, we observed that activations and weights of most DNNs do not follow a uniform distribution, which causes a huge impact in terms of quantization error and accuracy loss when the precision is further reduced to lower bitwidths. Specially in recent DNNs that are extremely deep and can have hundreds of layers, the error is propagated and expanded among layers. To determine the best quantization function, we first perform a goodness-of-fit comparison of different distributions over the tensors of a variety of layers from multiple popular DNNs. The Residual Sum of Squares (RSS) metric is used to measure the discrepancy between the tensor distributions and an estimated model. RSS is computed by Equation~\ref{eqn:rss}, where $y_{i}$ is the $i^{th}$ value of the variable to be predicted, $x_{i}$ is the $i^{th}$ value of the explanatory variable, and $f(x_{i})$ is the predicted value of $y_{i}$. In the analysis below, we use the absolute values of each tensor to measure the RSS.

\vskip -0.15in
\begin{equation}
\label{eqn:rss}
RSS = \sum_{i=1}^{n} (y_{i} - f(x_{i}))^2
\end{equation}

\begin{table}[t!]
\scriptsize
\caption{Mean RSS of activations for different distributions.}
\begin{center}
\vskip -0.15in
\begin{tabular}{|c|cccc|}
    \hline
    \textbf{DNN/Dist(RSS)} & \textbf{Normal} & \textbf{Exponential} & \textbf{Pareto} & \textbf{Uniform}\\
    \hline
    \textbf{Transformer} & 17.71 & \textcolor{red}{\textbf{2.82}} & 10.84 & 59.54 \\
    \textbf{ResNet-50} & 2.72 & \textcolor{red}{\textbf{0.71}} & 2.06 & 4.35 \\
    \textbf{AlexNet} & 15.81 & \textcolor{red}{\textbf{3.66}} & 22.07 & 23.14 \\
    \hline
\end{tabular}
\label{t:act_rss}
\end{center}
\vskip -0.15in
\end{table}

\begin{figure}[t!]
    \centering
    \begin{subfigure}[b]{0.42\columnwidth}
        \centering
        \includegraphics[width=\columnwidth]{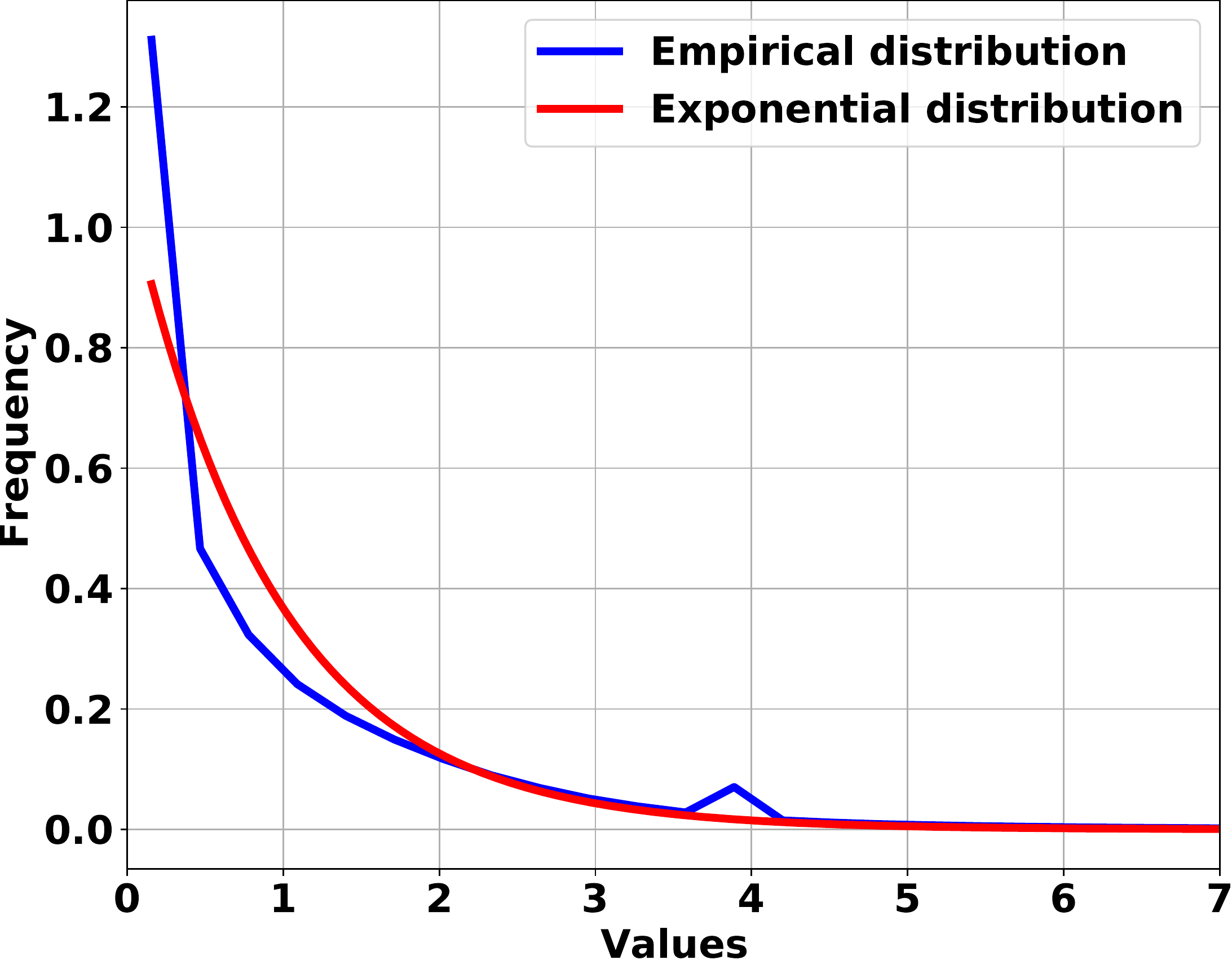}
        \caption{AlexNet CONV2 Activations}
        \label{fig:alexnet_l2_act}
    \end{subfigure}
    \begin{subfigure}[b]{0.42\columnwidth}
        \centering
        \includegraphics[width=\columnwidth]{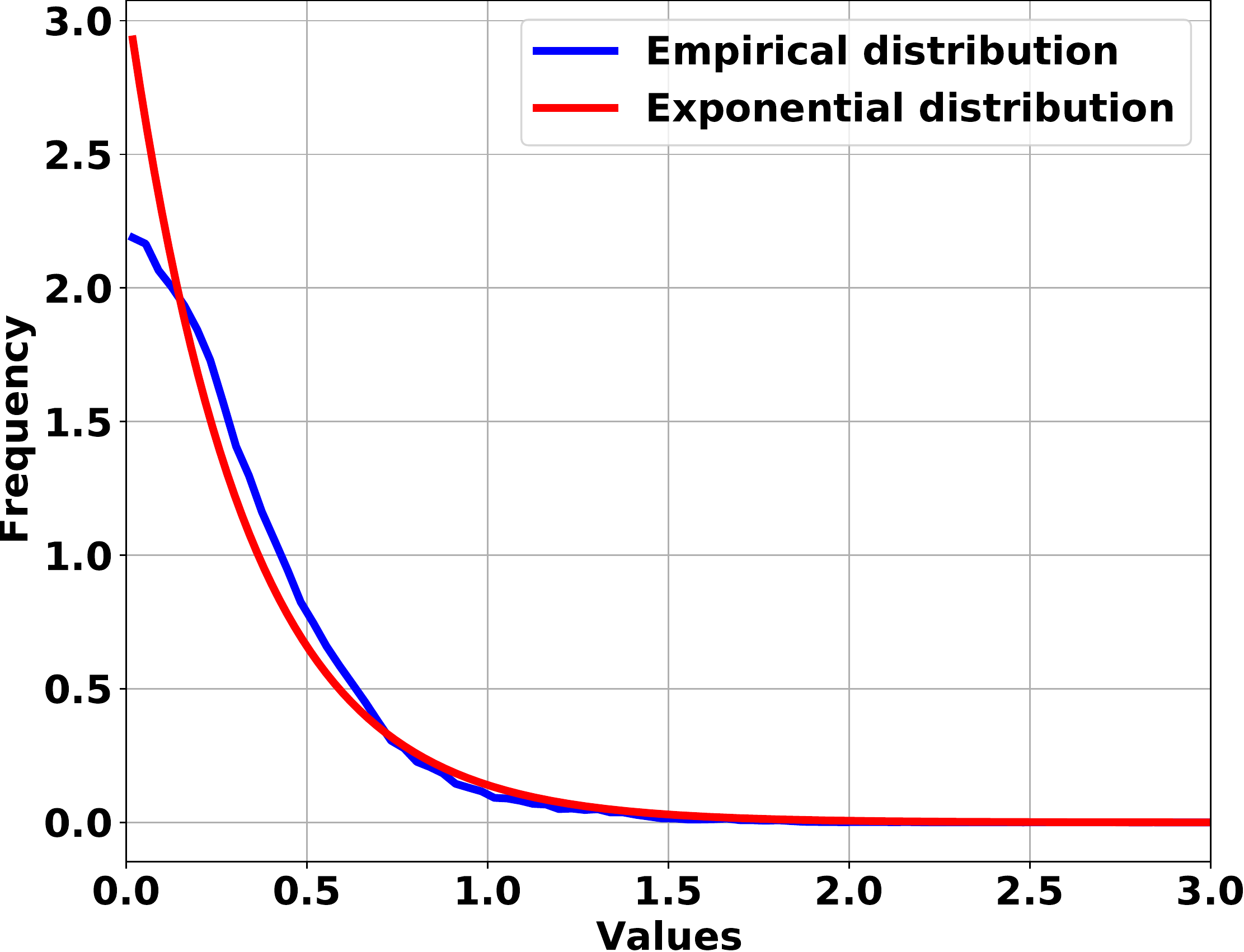}
        \caption{Transformer FC4 Activations}
        \label{fig:transformer_l4_act}
    \end{subfigure}
    \caption{Example of an exponential curve fitting to the empirical tensor of activations of a layer of (a) AlexNet and (b) Transformer DNNs.}
    \label{fig:act_curves_example}
\end{figure}

\begin{table}[t!]
\scriptsize
\caption{Mean RSS of weights for different distributions.}
\begin{center}
\vskip -0.15in
\begin{tabular}{|c|cccc|}
    \hline
    \textbf{DNN/Dist(RSS)} & \textbf{Normal} & \textbf{Exponential} & \textbf{Pareto} & \textbf{Uniform}\\
    \hline
    \textbf{Transformer} & 29.72 & \textcolor{red}{\textbf{9.86}} & 33.99 & 157.21 \\
    \textbf{ResNet-50} & 1667.32 & \textcolor{red}{\textbf{61.50}} & 142.38 & 4615.42 \\
    \textbf{AlexNet} & 1177.19 & \textcolor{red}{\textbf{179.79}} & 435.77 & 5591.18  \\
    \hline
\end{tabular}
\label{t:w_rss}
\end{center}
\vskip -0.15in
\end{table}

\begin{figure}[t!]
    \centering
    \begin{subfigure}[b]{0.42\columnwidth}
        \centering
        \includegraphics[width=\columnwidth]{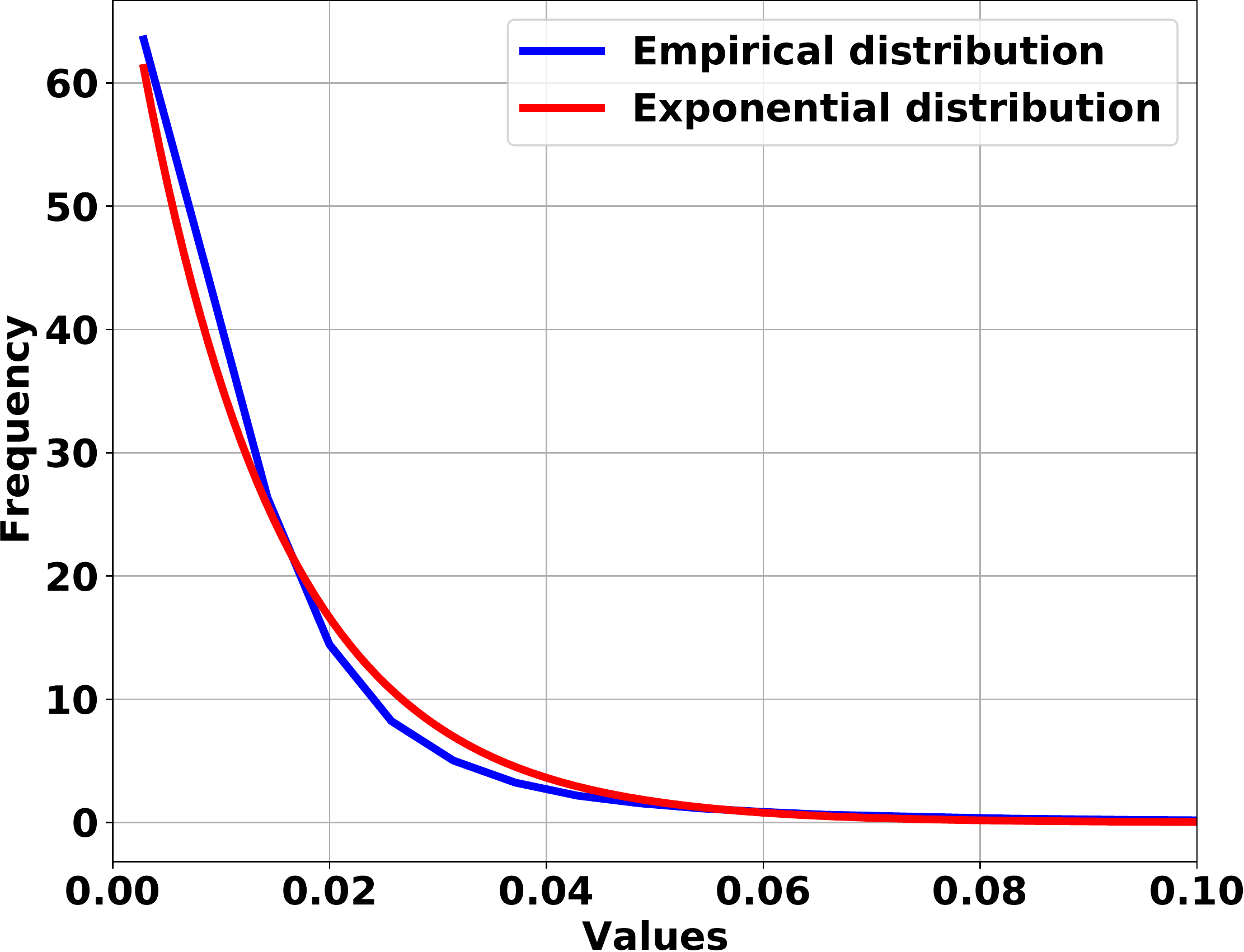}
        \caption{AlexNet CONV2 Weights}
        \label{fig:alexnet_l2_weights}
    \end{subfigure}
    \begin{subfigure}[b]{0.42\columnwidth}
        \centering
        \includegraphics[width=\columnwidth]{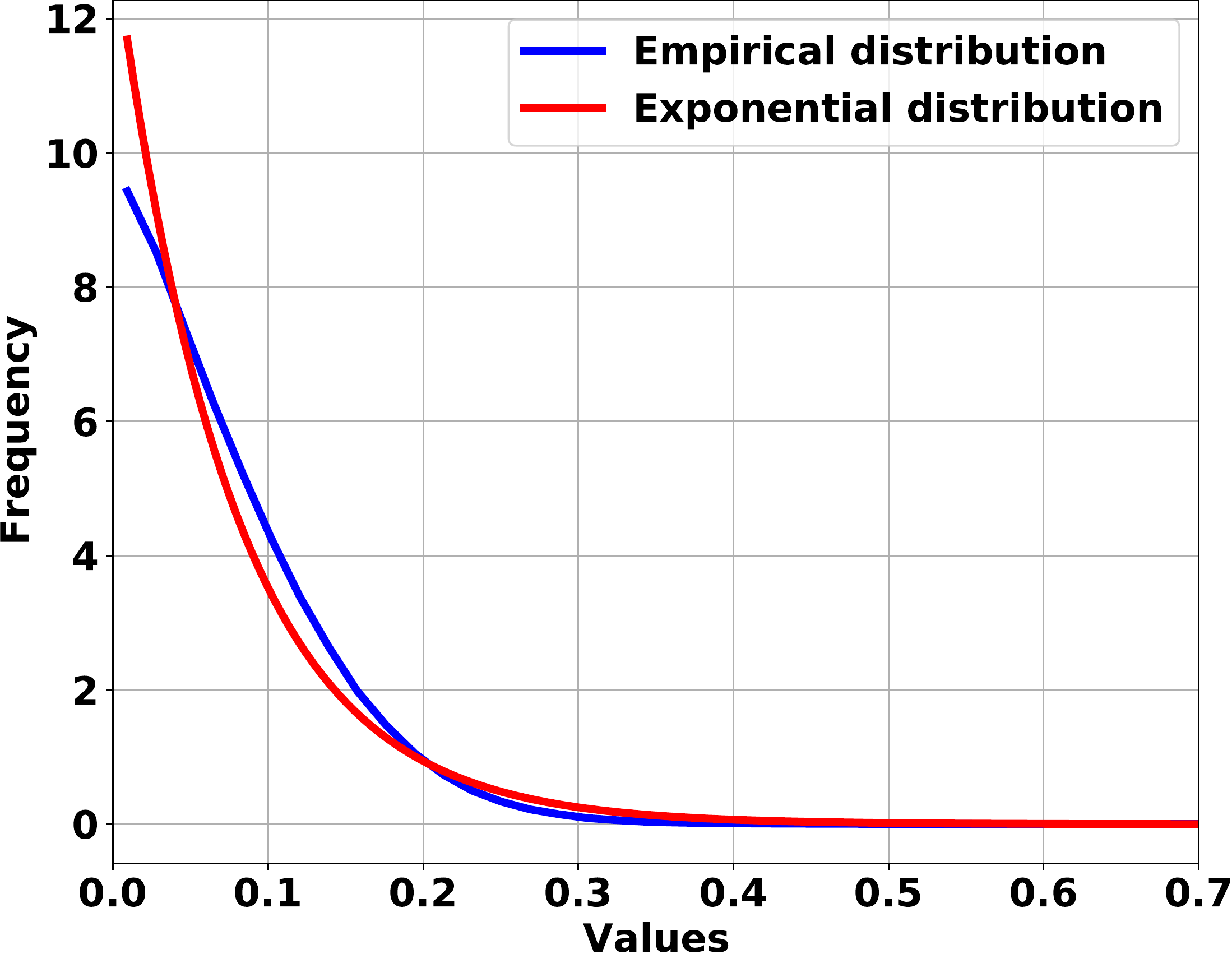}
        \caption{Transformer FC4 Weights}
        \label{fig:transformer_l4_weights}
    \end{subfigure}
    \caption{Example of an exponential curve fitting to the empirical tensor of weights of a layer of (a) AlexNet and (b) Transformer DNNs.}
    \label{fig:w_curves_example}
    \vskip -0.15in
\end{figure}

Table~\ref{t:act_rss} shows a comparison of the mean RSS of a set of common distributions for a trace of activations of all the FC and CONV layers of three different DNNs. For a given distribution and DNN, we report the mean RSS of all layers. As can be seen, the exponential distribution exhibits the lowest mean RSS, marked in red, for all the networks and, hence, an exponential curve is the closest fit for the majority of tensors of activations. As an example, Figure~\ref{fig:act_curves_example} plots the histograms and exponential curve fittings of the empirical activations for the CONV2 layer of AlexNet and the FC4 layer of the Transformer DNN, which results in RSS values of $0.02$ and $0.58$, respectively.

Similarly, Table~\ref{t:w_rss} shows the mean RSS of the four distributions for the weights of all layers of each DNN model. Again, the exponential distribution exhibits the lowest RSS compared to the other distributions. To illustrate it, Figure~\ref{fig:w_curves_example} depicts the histograms and plot fittings of the exponential curves of the empirical weights for the CONV2 layer of AlexNet and the FC4 layer of the Transformer DNN, which results in RSS values of $30.57$ and $3.4$, respectively. Compared to the activations, the RSS values of the weights are much larger. However, the exponential curve is still the best fit. In particular, the Transformer DNN exhibits the lowest RSS.

To summarize, we observe that the distributions of weights and activations, a.k.a. tensors, of a set of DNNs present a strong resemblance to exponential functions as demonstrated by its low RSS. Our aim is to quantize the tensors of all convolutional (CONV) and fully-connected (FC) layers with the nearest exponential representation that minimizes the quantization error, while achieving the best trade-off between numerical precision and hardware complexity. Next section describes how to obtain the pseudo-optimal parameters that define the closest-fit exponential function for each DNN layer and tensor.

\subsection{Searching Pseudo-Optimal Quantization Parameters}\label{search_alg}
In this section, we describe our adaptive methodology for searching offline the pseudo-optimal exponential quantization parameters on a layer-by-layer basis for any DNN. Based on the observations from Section~\ref{distribution_analysis}, we propose a quantization scheme to represent the activations and weights of DNNs in the form of an exponential function \(\bar{x} = Sign(x)(\alpha b^i + \beta\)), where $\bar{x}$ is the approximated value after the exponential quantization, $x$ is the original floating-point value of the tensor element, $b$ is the base of the exponential, $i$ is an integer exponent, $\alpha$ is an scale factor, and $\beta$ is an offset. The $Sign(x)$ function returns $0$ for zero values and $1/-1$ for positive and negative values. The integer exponent $i$ of each element $x$ in a tensor is computed with a logarithmic operation according to the following equations:

\vskip -0.05in
\begin{equation}
\begin{multlined}
\label{eqn:log_quant}
    LogExpQuant(x, b, \alpha, \beta, n) = \\
    Clip (Round(log_{b} (\frac{|x|- \beta}{\alpha})), -(2 ^ {n-1} - 1), (2 ^ {n-1} -1)),
\end{multlined}
\end{equation}

where

\vskip -0.05in
\begin{equation}
\label{eqn:clip}
    Clip(i, R_{min}, R_{max})= 
    \begin{cases}
        R_{min} & i\leqslant R_{min}\\
        R_{max} & i\geqslant R_{max}\\
        i & \text{otherwise}.
    \end{cases}
\end{equation}

In Equation~\ref{eqn:log_quant}, the $Round$ function is defined as the rounding to the nearest integer, $n$ is the number of bits required to represent the exponent $i$, and the clipping function in Equation~\ref{eqn:clip} forces the exponent values to be in the range of \([R_{min}, R_{max}]\), where $R_{min} = -(2^{n-1}-1)$ and $R_{max} = (2^{n-1}-1)$. Assuming an n-bit exponential quantization, the number of unique intervals is $2^{n}-1$. The approximated tensor values $\bar{x}$ are stored as exponents to reduce memory footprint since all other parameters are constants. We store an extra bit for the sign of the value, while the exponent $-(2^{n-1})$ is used as a special case to represent the zero value. On the other hand, the base $b$, bitwidth $n$, scale $\alpha$, and offset $\beta$ are constant parameters that are defined offline using an iterative algorithm that minimizes the quantization error and accuracy loss with the lowest possible number of intervals. For a given layer, we constrain $n$ and $b$ to be the same for both activations and weights in order to simplify the dot-product operations and the related hardware, by exploiting the property of exponentials: $b^i \times b^j = b ^{i+j}$. Note that the quantization parameters, including $n$ and $b$, can be different per layer and DNN to better fit the distributions of the corresponding tensors.

\begin{figure}[t!]
\centering
\includegraphics[width=0.82\columnwidth]{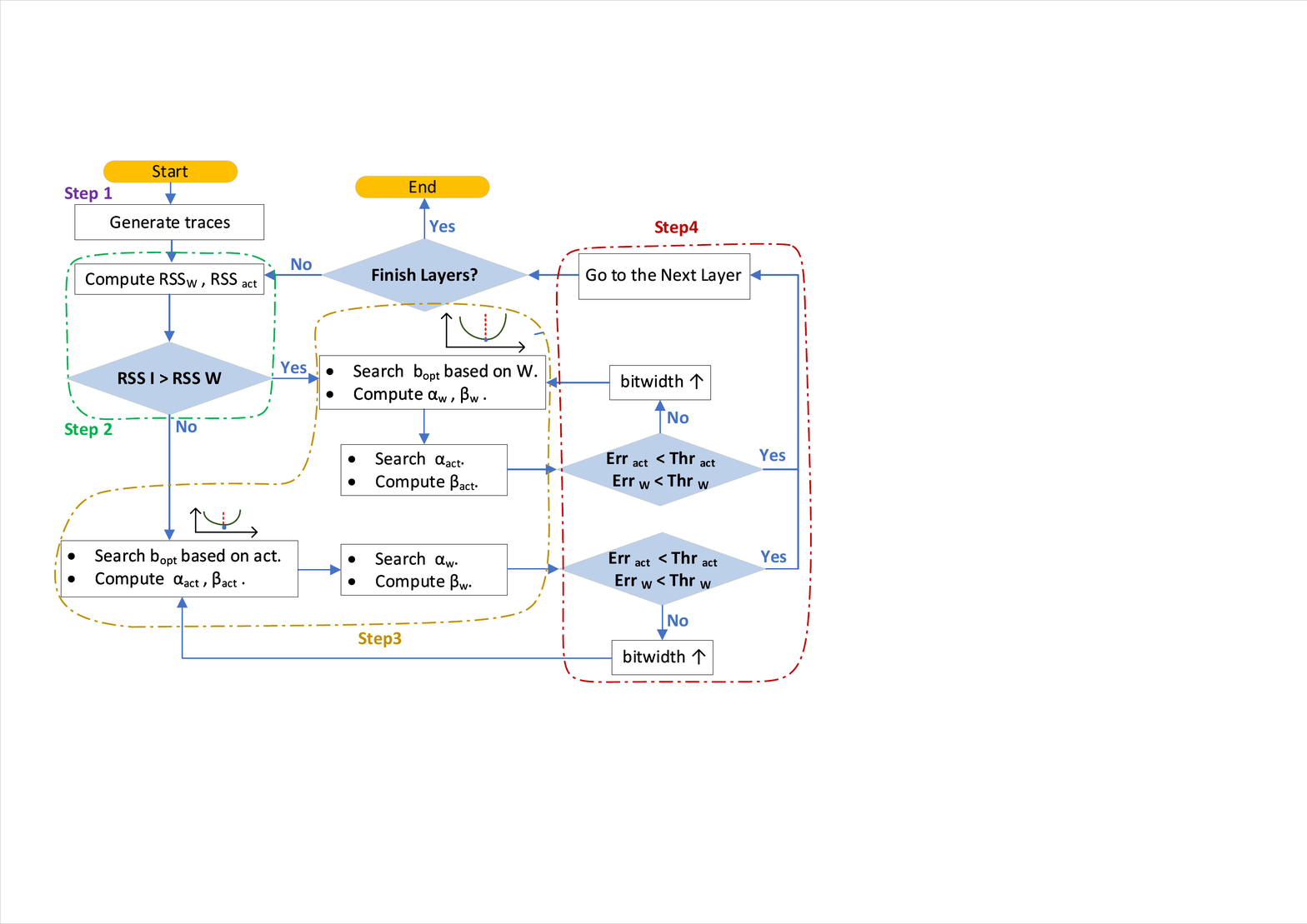}
\caption{DNA-TEQ Flowchart.}
\label{fig:flowchart}
\vskip -0.20in
\end{figure}

The flowchart in Figure~\ref{fig:flowchart} summarizes the four main steps of our proposed quantization methodology, DNA-TEQ. Each step is marked with a different color. Below, we describe in detail each of the steps to find the pseudo-optimal exponential quantization parameters that attain the best trade-off among accuracy, compression and hardware complexity. In the first step, we generate traces for the activations and weights of each layer to be quantized. For a given DNN, all the weights are obtained from the pre-trained model, while the trace of activations is generated from executing the inference of a small statistically representative subset of the training dataset.

The second step of DNA-TEQ, marked in green in Figure~\ref{fig:flowchart}, performs the computation of the RSS metric of the tensors of each layer based on Equation~\ref{eqn:rss}. For a given layer, the tensor with the smaller RSS is selected to start the search of the base $b$. In all the layers of AlexNet and ResNet-50, the tensor of activations is chosen for computing the base, whereas in the Transformer DNN, 12 out of 96 FC layers choose the tensor of weights. The goal of this step is to start the search of the pseudo-optimal base from the tensor that has more similarity to the exponential distribution, reducing the induced error.

The third step, marked in yellow in Figure~\ref{fig:flowchart}, starts the offline search for the pseudo-optimal base and the rest of the parameters of the exponential quantization for a specific layer $l$. As described above, the initial tensor $t$ is selected from either weights or activations depending on the RSS metric. The base $b_{l}$ and scale $\alpha_{lt}$ are initialized taking into account the maximum value of the tensor to cover the full scale range (FSR) with the exponential quantization as shown by Equation~\ref{eqn:base_scale}. Our empirical experiments demonstrated that by covering the FSR of the tensor, the quantization error is reduced due to the effect of large outliers. In addition, FSR helps to find the pseudo-optimal point of convergence of the search algorithm faster. On the other hand, the distribution analysis of activations and weights shows that most values are clogged in a range close to the minimum value of the tensor. Therefore, the offset $\beta_{lt}$ is initialized so that the smallest values of the tensor can be represented more precisely. In Equation~\ref{eqn:offset}, the first term shifts the quantization intervals close to the tensor minimum value, while the second term takes into account the rounding effect from Equation~\ref{eqn:log_quant}.

\vspace*{-.1cm}
\begin{equation}
\label{eqn:base_scale}
b_{l} = max(t)^\frac{1}{R_{max}} ; \alpha_{lt} = \frac{max(t)}{b ^ {R_{max}}}
\end{equation}

\begin{equation}
\label{eqn:offset}
\begin{multlined}
\beta_{lt} = \underbrace{min(t) - \alpha_{lt} b ^{R_{min}}}_\text{1} + \underbrace{\alpha_{lt} b ^{R_{min}} - \alpha_{lt} b ^{R_{min} - 0.5}}_\text{2}
\end{multlined}
\end{equation}
\vspace*{-.2cm}

For a given tensor, Algorithm~\ref{alg:sob} shows the pseudo-code for searching the optimal base and the related exponential quantization parameters that provide the lowest quantization error. First, we initialize the base $b$, scale $\alpha$, and offset $\beta$ with the equations described above (line 2). Then, we perform the initial quantization and compute $InitErr$ (line 3) using the Relative Mean Absolute Error (RMAE) metric as defined in Equation~\ref{eqn:MAE}. Next, we decide if the exploration of the base is going to be done by increasing or decreasing the initial base (lines 4-8) by a delta $\varepsilon$ that is initialized to $0.01$. The actual direction is selected by taking into account the lowest error among $InitErr$, $IncErr$ and $DecErr$. Before computing the errors, the corresponding $\alpha$ and $\beta$ are calculated for $IncBase/DecBase$. Finally, we continue to search for the pseudo-optimal base (lines 9-19) by repeatedly increasing/decreasing the base until the quantization error $CurrentErr$ is no longer reduced. In the process, we update the corresponding scale $\alpha$ and offset $\beta$ taking into account the $NewBase$ according to Equations~\ref{eqn:base_scale}-\ref{eqn:offset}. Once the parameters of a tensor (activations or weights) of a layer are computed, for the other tensor of this layer the same base is used, and we simply compute the $\alpha$ and $\beta$ parameters in the same manner as for the other tensor.



\begin{algorithm}[t!]
\scriptsize 
\caption{Searching Pseudo-Optimal Base}
\label{alg:sob}
\begin{algorithmic}[1]
    \Procedure{SOB}{$t, b, n,\alpha,\beta$}\Comment{in:$t$, $n$}\Comment{out:$b,\alpha,\beta$}
    \State Initialize($b, \alpha, \beta$);
    \State $InitErr = RMAE(LogExpQuant(t, b, \alpha, \beta, n), t)$;
    \State $\varepsilon = 0.01$;
    \State $IncBase = b + \varepsilon$; $DecBase = b - \varepsilon$;
    \State $Alpha[], Beta[] = Update(\alpha, \beta, IncBase, DecBase)$;
    \State $IncErr, DecErr = RMAE(t, n, IncBase, DecBase, Alpha[], Beta[])$;
    \State $CurrentErr, \varepsilon, b = Direction(InitErr, IncErr, DecErr)$;
    \While{$Search = True$}
        \State $NewBase = b + \varepsilon$;
        \State $NewAlpha, NewBeta = Update(\alpha, \beta, NewBase)$;
        \State $NewErr = RMAE(t, n, NewBase, NewAlpha, NewBeta)$;
        \If{$NewErr < CurrentErr$}
            \State $CurrentErr = NewErr$;
            \State $b = NewBase$; $\alpha = NewAlpha$; $\beta = NewBeta$;
        \Else 
            \State $Search = False$
        \EndIf
    \EndWhile
    \EndProcedure
\end{algorithmic}
\end{algorithm}
\vspace*{-.2cm}

\begin{equation}
\label{eqn:MAE}
RMAE = \frac{\sum |\Bar{t} - t|}{\sum |t|} 
\end{equation}

In the fourth step, marked in red in Figure~\ref{fig:flowchart}, and for a given layer, DNA-TEQ iterates over the quantization bitwidth $n$ starting from 3 bits until a maximum of 7 bits, to find the lowest bitwidth that does not hurt accuracy. For all layers of a DNN, an error threshold is defined for each tensor, $Thr_{act}$ for activations and $Thr_{w}$ for weights, as the maximum quantization error (RMAE). $Thr_{w}$ is initialize to 1\% while $Thr_{act}$ is scaled by Equation~\ref{eqn:inputs_threshold} to account for the difference in magnitude between the two distributions. Once the quantization parameters for a given bitwidth are obtained, the errors are compared against the thresholds, repeating the search algorithm until the condition is fulfilled.

\vskip -0.20in
\begin{equation}
\label{eqn:inputs_threshold}
\ Thr_{act} = Thr_{w} \times log(\frac{mean(Act.)}{mean(W)})
\end{equation}



After all layers have obtained their respective quantization parameters, we perform DNN inference to check accuracy loss, and iterate over $Thr_{w}$ in steps of 1\%, that is, if the accuracy loss is negligible we increase the error that can be tolerated in each layer and re-run the search of parameters. This procedure is continued while the accuracy loss is lower than 1\%. In the evaluation section, we show a sensitivity analysis of $Thr_{w}$.

\subsection{Exponential Dot-Product}\label{qt_error_hardware}
As previously discussed, our aim is to exploit the quantization not only to reduce memory footprint but also hardware complexity. In this section, we will first exploit that values are quantized with an exponential function to simplify the dot-product operation, replacing the costly multiplications by simple additions. Then, we provide some hints on the requirements to implement and take advantage of our quantization scheme, including an initial hardware configuration of DNA-TEQ.

At a high-level, DNA-TEQ takes advantage of the property $b^{a} \cdot b^{w} = b^{a + w}$. After quantizing the values, all values are of the form $S_{W}(\alpha_{W} b^{int_{W}} + \beta_{W})$ and $S_{A}(\alpha_{A} b^{int_{A}} + \beta_{A})$ for weights $W$ and activations $A$ respectively, where $S_{W}$, $S_{A}$ are signs, $int_{W}$, $int_{A}$ are signed $n$-bit integer exponents, $\alpha_{W}$, $\alpha_{A}$ are scale factors and $\beta_{W}$, $\beta_{A}$ are offsets for the corresponding weight and activation tensors, respectively. As shown in Equation~\ref{eqn:conv_extended}, each output activation is a sum of activations times weight products, which can be expanded into a sum of four terms:

\vskip -0.15in
\begin{equation}
\begin{small}
\begin{multlined}
\label{eqn:conv_extended}
\sum_{i=1}^{m} A_{i} \cdot W_{i} = \sum_{i=1}^{m}S_{A_{i}} (\alpha_{A} b^{int_{A_{i}}} + \beta_{A}) \cdot S_{W_{i}} (\alpha_{W} b^{int_{W_{i}}} + \beta_{W}) =\\
\underbrace{\alpha_{A}\alpha_{W}\sum_{i=1}^{m} (S_{A_{i}} S_{W_{i}})b^{int_{A_{i}} + int_{W_{i}}}}_\text{1} + \underbrace{\alpha_{W}\beta_{A}\sum_{i=1}^{m}(S_{A_{i}} S_{W_{i}})b^{int_{W_{i}}}}_\text{2} \\
+ \underbrace{\alpha_{A}\beta_{W}\sum_{i=1}^{m}( S_{A_{i}} S_{W_{i}}) b^{int_{A_{i}}}}_\text{3} + \underbrace{\beta_{A}\beta_{W}\sum_{i=1}^{m}S_{A_{i}} S_{W_{i}}}_\text{4}
\end{multlined}
\end{small}
\end{equation}

The first term is the sum of $b^{int_{A_{i}} + int_{W_{i}}}$, which can be implemented with a table of $2^{n+1}$ entries, where each entry stores the count of how many times each addition of exponents occurs. The sign bit of activation and weight are XORed and based on the result, the corresponding entry in the table is increased/decreased when $S_{A_{i}} S_{W_{i}}$ becomes positive/negative. The second term is the summation of weights with respect to the sign of both weight and activation, while the third term is similar but the summation of activations. Both can be computed like the first term with a table that only requires $2^{n}$ entries for each one due to the limited range of the exponent. The last term is the accumulation of the sign products which can be obtained from any of the previous terms by adding the total number of occurrences. Note that the second and fourth terms depend exclusively on the weights, which are known, as well as the signs of the activations, which are mostly positive due to the ReLU activation function. Therefore, these two terms can be pre-computed offline for the majority of DNN layers. After filling the tables, we multiply each count with its corresponding value ($b^{int}$) and accumulate all products into a single value. The final values are multiplied by the constant coefficients and all terms are added together producing the final output activation.


\section{Software Implementation}\label{simd_implementation}
In this section, we propose a software implementation of DNA-TEQ using Intel SIMD extensions that leverage the vector processing capability of CPUs. Then, we discuss the scalability issues and limitations of SIMD when implementing DNA-TEQ compared to an optimized SIMD version of INT8.

\begin{figure}[t!]
\centering
\includegraphics[width=0.9\columnwidth]{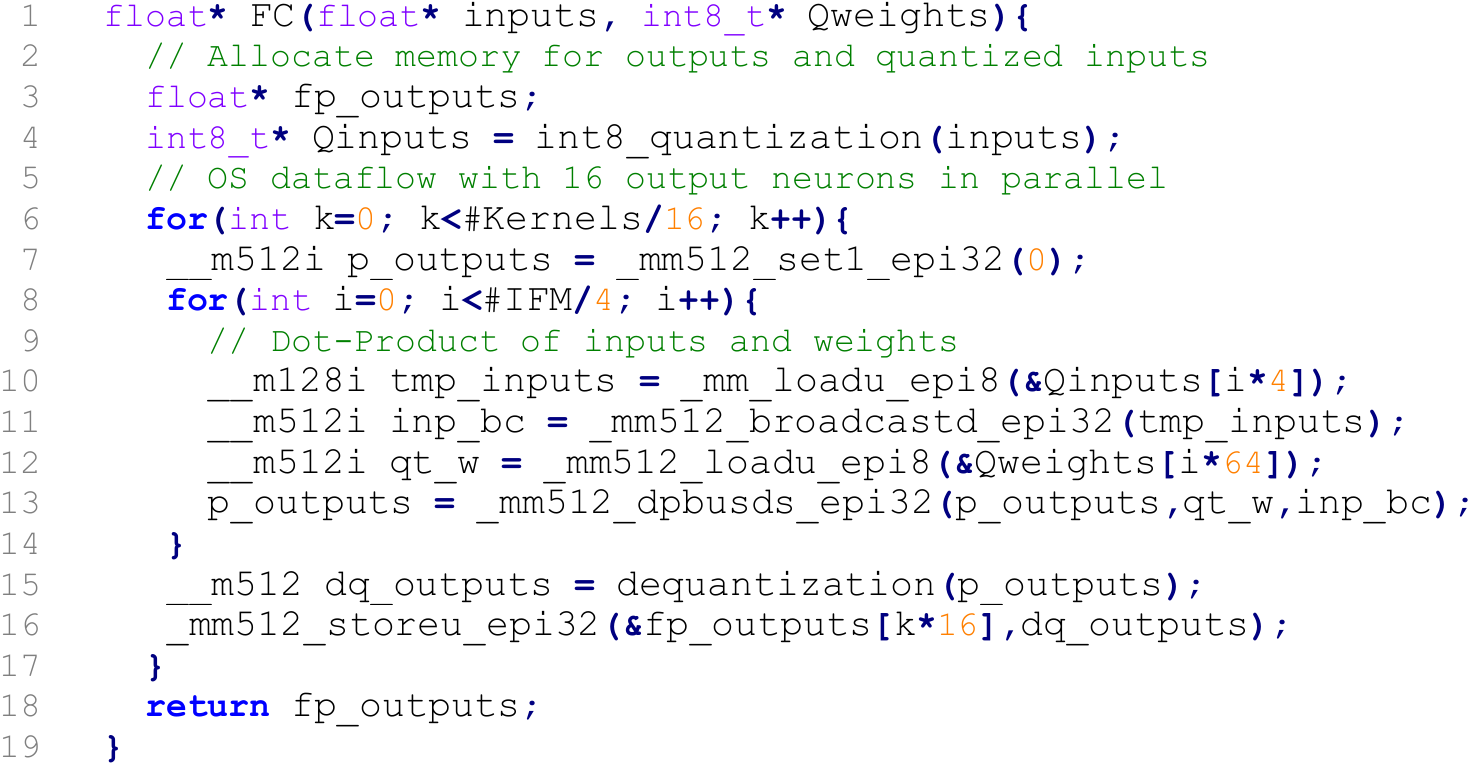}
\vskip -0.05in
\caption{Pseudo-code of the best-effort FC INT8 SIMD implementation.}
\label{fig:int8_simd}
\vskip -0.20in
\end{figure}

To efficiently implement the popular INT8 DNN models, Intel introduced the Vector Neural Network Instructions (VNNI)~\cite{VNNI} as an extension of the AVX-512 set. One of the key instructions, $VPDPBUSD$, fuses three operations into a single one, accelerating the inner-loop multiply-accumulate of INT8 convolutions. This instruction can simultaneously perform 4 MAC operations for 16 different output neurons. Figure~\ref{fig:int8_simd} presents the pseudo-code of our best-effort to implement an FC layer with INT8 VNNI. Quantization and de-quantization functions are also implemented using SIMD intrinsics, although not shown in the code for the sake of simplicity.

\begin{figure}[t!]
\centering
\includegraphics[width=0.9\columnwidth]{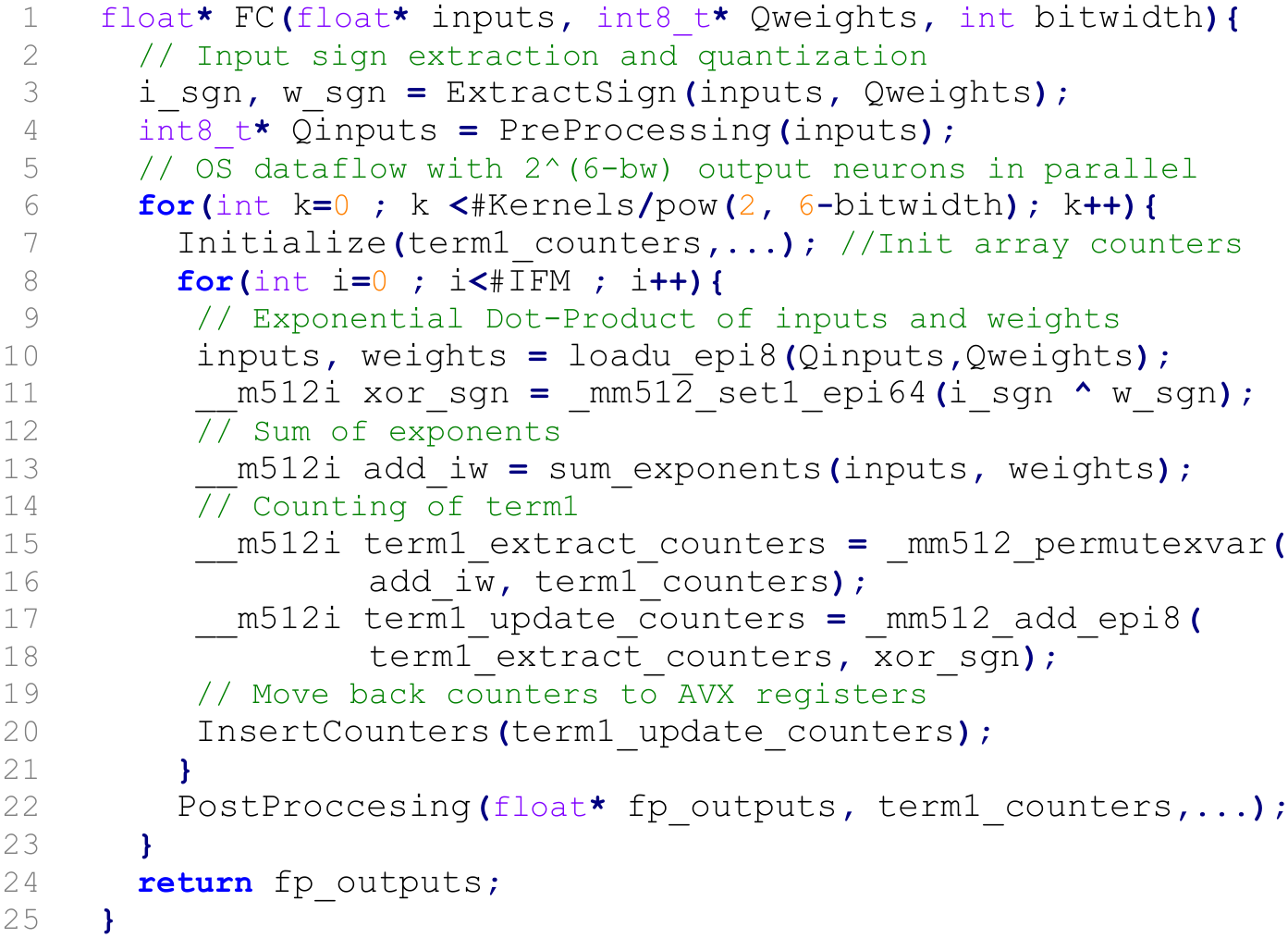}
\vskip -0.05in
\caption{Psuedo-code of the DNA-TEQ FC SIMD implementation.}
\label{fig:dna_teq_simd}
\vskip -0.05in
\end{figure}

Figure~\ref{fig:dna_teq_simd} depicts the pseudo-code of the DNA-TEQ SIMD implementation of an FC layer. As described in Section~\ref{qt_error_hardware}, DNA-TEQ performs the dot-product by counting the frequency of exponents for each term. Therefore, each layer requires a different amount of 8-bit counters per neuron (i.e. array counters) based on the numerical precision determined by our search algorithm. These array counters can be allocated in memory, and the Scatter-Gather instructions can read from or write to them based on the pointers. However, the major drawback of Scatter-Gather is that it produces a huge amount of data movements in the memory hierarchy, incurring in a significant latency per operation. To mitigate this issue, we allocate the array counters within SIMD registers. These registers are limited both in quantity (e.g. 32) and size (512 bits), reducing the amount of indexations (i.e. counting operations) that can be done in parallel, since we can only index one register at a time. For example, the level of parallelism is restricted to 2 and 4 different output neurons when the numerical precision is 5-bit and 4-bit, respectively. In the best case scenario, when the numerical precision is 3-bit, 8 output neurons can be computed concurrently. Intel also provides with the $permutexvar$ instruction, which we use to shuffle the registers of the array counters using the corresponding exponents as indexes.


Table~\ref{t:simd_exec_time} presents the execution times of FC layers for different configurations and schemes. DNA-TEQ can provide significant speedup over the baseline and, in particular, up to $5x$ when the size of the FC layer becomes bigger. This is mainly due to the replacement of multiplications for counting operations and the reduction of data movements. DNA-TEQ operates within SIMD registers avoiding expensive load/store. However, we observe a slowdown when the precision starts to increase. This can be attributed to the number of counters required and the limited amount of SIMD registers and its size. In the worst case, when precision is higher than 6-bit, we can not allocate enough registers, and need to relay on additional transformations to perform the counting, incurring in more data movements. On the other hand, VNNI is highly optimized for the INT8 representation, providing a substantial degree of parallelism. Consequently, dedicated hardware is required to exploit all the advantages offered by DNA-TEQ.


\begin{table}[t!]
\scriptsize 
\caption{Execution time (in milliseconds) comparison of FC layers of varying sizes implemented using SIMD for INT8 and DNA-TEQ. Experiments executed using a Xeon W-2245 CPU.}
\begin{center}
\vskip -0.10in
\begin{tabular}{|c|ccc|}
    \hline
    \textbf{Quantization Scheme} & \textbf{FC(1024, 1024)} & \textbf{FC(2048, 2048)}& \textbf{FC(4096, 4096)} \\
    \hline
    \textbf{Uniform INT8} & 0.11 & 0.37 & 5.66 \\
    \hline
    \textbf{DNA-TEQ 3-bit}& 0.17 & 0.35 & 1.11 \\
    \textbf{DNA-TEQ 4-bit}& 0.34 & 0.88 & 2.14 \\
    \hline
\end{tabular}
\label{t:simd_exec_time}
\end{center}
\vskip -0.20in
\end{table}

\section{Hardware Implementation}\label{accelerator_architecture}
This section outlines the hardware support required for implementing DNA-TEQ. We begin by introducing the essential hardware components of the DNA-TEQ accelerator. Subsequently, we provide a detailed description of the execution process of DNN layers within the accelerator divided into three stages: 1) Pre-Processing, 2) Counting, and 3) Post-Processing.



\begin{figure}[t!]
\centering
\includegraphics[width=0.9\columnwidth]{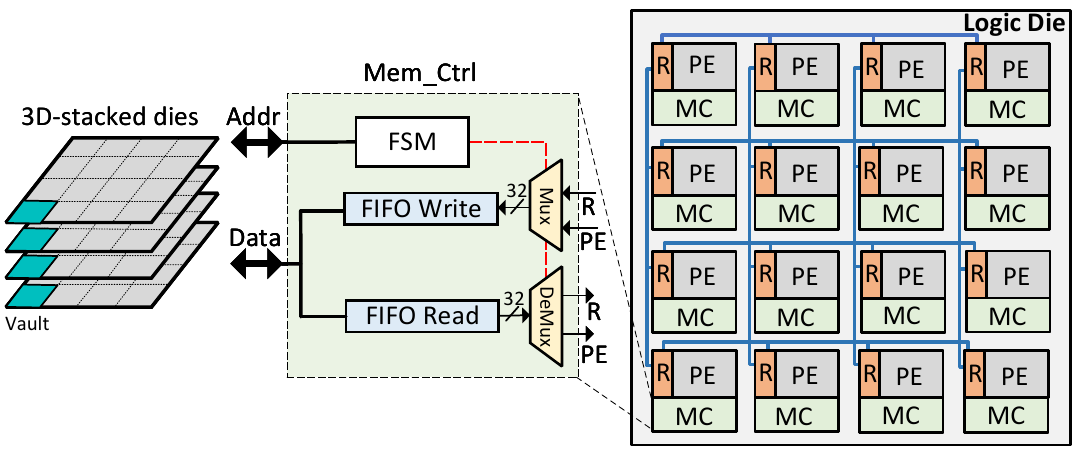}
\vskip -0.05in
\caption{Overview of the DNA-TEQ Accelerator.}
\label{fig:overview_hw}
\vskip -0.20in
\end{figure}

\begin{figure*}[t!]
\centering
\includegraphics[width=0.9\textwidth]{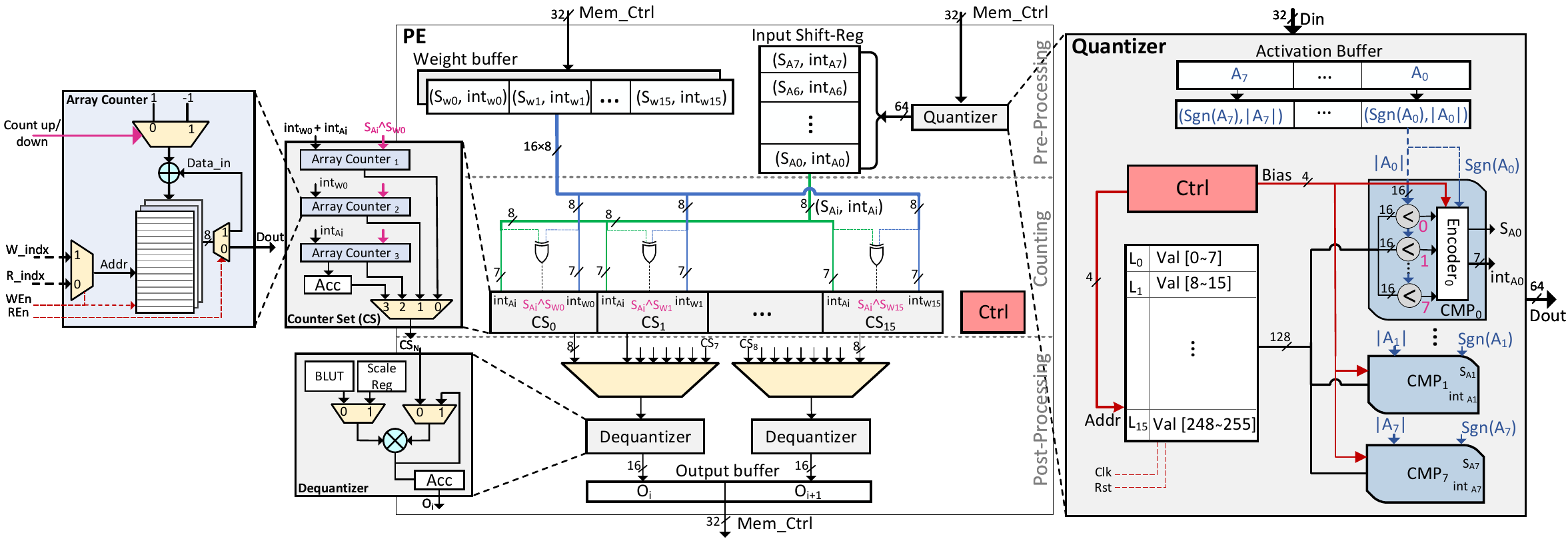}
\caption{Architecture of a Processing Element (PE) including the organization of a Quantizer, Dequantizer, Counter Set (CS), and Array Counter.}
\label{fig:pe_hw}
\vskip -0.20in
\end{figure*}

\subsection{Accelerator Architecture}
Figure~\ref{fig:overview_hw} presents an overview of the accelerator architecture, which relies on 3D-stacked memory for storing activations and weights. Within the logic die, each tile comprises a single Processing Element (PE), a Memory Controller (MC), and a Router (R). The MC manages all memory operations within the associated vaults in the DRAM dies. Additionally, the MC incorporates two FIFOs to facilitate multi-domain frequency synchronization between the logic die and the DRAM dies. The Router enables local and remote accesses between PEs and vaults via a 2D mesh network. Finally, the PE is the core of the tile, and is responsible for accelerating the DNN operations by computing output activations based on Equation~\ref{eqn:conv_extended}. Figure~\ref{fig:pe_hw} shows the main components of a PE divided into three main stages. Below is a detailed description of each stage.

\subsection{Pre-Processing Stage}
In the first stage, DNA-TEQ performs exponential quantization on the activations to extract their signs ($S_{A}$) and exponents ($int_{A}$) at runtime, in batches of eight. On the other hand, all weights are pre-quantized offline without requiring additional hardware. The $Quantizer$ unit is in charge of mapping the activations from floating-point (FP16) to the nearest integer exponents according to the numerical precision of each layer.

During the quantization process, each activation is assigned to a CMP module ($CMP_{0}$ to $CMP_{7}$) to compare against all the boundaries of the exponential quantization intervals. These ranges are loaded during the execution of each layer and stored in a multi-banked memory buffer ($L_{0}$ to $L_{15}$) sized with a total of 256B for the worst-case scenario (i.e. 7-bit exponents). Comparisons are also performed in batches of 8 values to conduct the 3-bit quantization of multiple activations in a single cycle. To this end, each CMP module includes a set of 8 comparators that select the proper values from each bank of the memory buffer. Since the boundaries are sorted, the output of the comparators is an 8-bit vector of 0s followed by 1s. Based on the comparison results, the encoder detects the leading one and selects the corresponding $int_{A}$ value for the activation. Note that this quantization unit is less costly than implementing a complex log-base module, and banks of memory that are not needed for a given layer can be power-gated. The resulting exponents and their corresponding signs ($S_{Ai}$, $int_{Ai}$) are stored in the Input Shift-Reg. This arrangement ensures that, in each cycle, a single quantized activation is fetched and broadcasted to all Counter-Sets ($CS_{0}$ to $CS_{n-1}$) to be used in the subsequent stage. At the same time, the Weight Buffer receives $N$ quantized weights and their signs ($S_{wi}, int_{wi}$) from the Memory Controller.


\subsection{Counting Stage}
The counting stage is the core of our accelerator and consists of a controller to orchestrate all the execution, and $N$ Counter-Sets ($CS_{i}$) to compute $N$ different output neurons at the same time (e.g. 16). Each Counter-Set is composed of several components, including an adder to perform the sum of exponents, an accumulator to account for the occurrences of sign products of the fourth term when needed, and three Array Counters ($AC_{1}$-$AC_{3}$) to calculate the other terms of Equation~\ref{eqn:conv_extended}. Specifically, $AC_{1}$ is devoted to counting the occurrences of exponents required for computing the first term, while $AC_{2}$ and $AC_{3}$ are used for the second and third terms, respectively.

As shown in the left of Figure~\ref{fig:pe_hw}, each $AC$ includes an adder, two multiplexers, and a small multi-banked SRAM buffer. Each SRAM has a total of 16 banks. All these buffers are sized for the worst-case scenario (i.e. 7-bit exponents). In particular, $AC_{2}$ and $AC_{3}$ have 8 entries per bank with a total of 128B, while $AC_{1}$ has 16 entries per bank with a total of 256B. For our set of DNN benchmarks, 8-bit per entry is enough to perform the count of each term without any numerical instability. To minimize the static energy of $ACs$ among the multiple PEs, the unused banks will be power-gated according to the quantization bitwidth of each layer. For example, if the numerical precision of a layer is 3-bit, only one bank of each $AC$ will be active.

\subsection{Post-Processing Stage}
The final stage of the accelerator is composed of two de-quantization units, two multiplexers, and an Output buffer. Each Dequantizer consists of an FP16 multiplier, an accumulator, and two small buffers shared between de-quantization units: a Base-Lookup Table (BLUT) and a Scale Register.


During the post-processing, the $CSs$ retain the occurrences of exponents that were computed in the counting stage, and send them to a Dequantizer. Next, each count of the $ACs$ is multiplied by its corresponding $b^{int}$ value and accumulated to process each term in Equation~\ref{eqn:conv_extended}. The FP16 base $b$ powers are loaded and stored for every layer in a BLUT of $2^{n+1}$ entries. Then, each dot-product result is multiplied by their related constant coefficients in the Scale-Register. Lastly, all terms are added together to provide the final output activation of a neuron.

The counting and post-processing stages are performed serially, while the pre-processing is done concurrently to hide the latency of the quantization. Serial post-processing avoids flushing the counters to a temporal buffer and additional data movements, while its latency is very small compared to the counting stage. Therefore, one de-quantization unit per Counter-Set does not significantly improve the overall speed. On the other hand, the area overhead associated with the FP16 multipliers is huge. Thus, we only employ two de-quantization units per PE to achieve the best trade-off. This configuration strikes a balance between performance and area, allowing for efficient processing of the output activations.



%% file: sections/4-evaluation.tex
\section{Experimental Results}\label{evaluation}
This section evaluates our proposal in terms of accuracy, compression, performance and energy efficiency. First, we describe the methodology and tools employed to evaluate DNA-TEQ. Then, we introduce an analysis of the accuracy and compression of our quantization scheme across various DNNs. Next, we present the speedups and energy savings achieved by our hardware accelerator compared to the INT8 baseline. Furthermore, we discuss the accelerator overheads. Finally, we conduct a sensitivity analysis on the DNA-TEQ parameters.


\subsection{Methodology}
We have developed a simulator that accurately models two different systems, our DNA-TEQ accelerator and a baseline with uniform INT8 quantization. The baseline architecture draws inspiration from 3D-stacked DRAM-based DNN accelerators such as Neurocube~\cite{neurocube} and Tetris~\cite{tetris}. Both accelerators implement a similar architecture, including an output stationary dataflow, and hardware units to perform the quantization and de-quantization of input/output activations based on their respective schemes, that is, DNA-TEQ and INT8. Besides, weights are quantized offline in both accelerators. For a fair comparison, we set most of the configuration parameters to be the same for both the baseline and DNA-TEQ: a 3D-stacked memory of 4 GB with 4 DRAM dies partitioned into $4 \times 4$ vaults and PEs, an internal bandwidth of 10 GB/s per vault, about 2.5 KB of SRAM per PE to store inputs/outputs/weights, 16 MAC or Counter-Set units per PE, and a frequency of 300 MHz in the logic die. DNA-TEQ requires 6 KB more of on-chip memory per PE due to the Counter-Sets that replace the MACs.



Regarding area and energy consumption evaluation, the logic components are implemented in Verilog, including all the additional components required by DNA-TEQ, and synthesized to obtain the delay, area, and power using the Synopsys Design Compiler, the modules of the DesignWare library, and the technology library of 28/32nm from Synopsys. On the other hand, we characterize the memory buffers of the accelerator by obtaining the delay, energy per access and area using CACTI-P. We use the configurations optimized for low power and a supply voltage of 0.78V. Finally, the energy consumption of the 3D-stacked memory is estimated by using DRAMSim3. The results obtained with the aforementioned tools are combined with the activity factors and memory traces provided by our simulator to obtain the dynamic and static power of the accelerators.


We evaluate our method on the classification task of ImageNet (ILSVRC-2012) using pre-trained models of popular DNNs including AlexNet and ResNet-50. In addition, we also evaluate a Transformer model on the machine translation task of Newtest2014 (English-German) which contains 3003 sentences. Python and Tensorflow are used to implement DNA-TEQ and the DNN models to assess the accuracy and measure the RSS.


\subsection{DNA-TEQ Evaluation}
The algorithm described in Section~\ref{search_alg} finds the optimal bitwidth for each layer and their corresponding quantization error. Table~\ref{t:qt_cmp} shows a comparison between uniform quantization and DNA-TEQ in terms of accumulated RMAE of activations and weights among all layers. In addition, the table reports the accuracy loss of both quantization schemes compared to the FP32 baseline, assuming the same number of bits obtained from the search algorithm for both cases. As can be seen, DNA-TEQ provides the lowest error and loss in all the evaluated DNNs.

\begin{table}[t!]
\scriptsize 
\caption{Error/Loss comparison between quantization schemes.}
\begin{center}
\vskip -0.15in
\begin{tabular}{|c|ccc|}
    \hline
    \textbf{DNN (RMAE/Accuracy Loss)} & \textbf{AlexNet} & \textbf{ResNet-50}& \textbf{Transformer} \\
    \hline
    \textbf{Uniform Quantization} & 7.02/18.3\% & 34.16/65.41\% & 127.75/27.5 \\
    \textbf{DNA-TEQ (Ours)}& 1.80/0.97\% & 1.39/0.45\% & 34.87/0.82 \\
    \hline
\end{tabular}
\label{t:qt_cmp}
\end{center}
\vskip -0.15in
\end{table}

\begin{table}[t!]
\scriptsize 
\caption{Effect of DNA-TEQ on accuracy, average bitwidth and compression ratio for AlexNet, ResNet-50 and Transformer DNNs.}
\begin{center}
\vskip -0.15in
\begin{tabular}{|c|cccc|}
    \hline
    \multirow{2}{*}{\textbf{Network}} & \textbf{Baseline Acc.} & \textbf{DNA-TEQ} & \textbf{AVG} & \textbf{Compression} \\
                     & \textbf{(FP32/INT8)} & \textbf{Accuracy} & \textbf{Bitwidth} & \textbf{Ratio (\%)} \\
    \hline
    \textbf{Transformer} & 27.93/27.87 & 27.11(BLEU) & 3.05 & 61.86 \\
    \textbf{ResNet-50} & 76.76/76.44\% & 76.31\%(Top-1) & 5.65 & 29.26 \\
    \textbf{AlexNet} & 57.48/57.22\% & 56.51\%(Top-1) & 5.78 & 27.64 \\
    \hline
\end{tabular}
\label{t:avg_bw_accy}
\end{center}
\vskip -0.25in
\end{table}

Table~\ref{t:avg_bw_accy} shows the average quantization bitwidth achieved by DNA-TEQ. The table also reports the accuracy of the baseline with 8-bit uniform quantization and DNA-TEQ without re-training. On average, DNNs are quantized to 4.83-bits, resulting in a compression ratio of 40\% over the linear INT8 baseline. In all cases, the accuracy loss is less than 1\% with respect to the FP32 baseline. These results are well correlated with the observations from Section~\ref{distribution_analysis}, proving that exponential quantization can reduce numerical precision below 8 bits.

As previously discussed, most current works on quantization either require re-training or target a specific type of DNN model. However, re-training is expensive in terms of time, energy, and system resources, hence our aim is to perform quantization without re-training and with negligible accuracy loss. Nevertheless, note that by doing re-training on top of DNA-TEQ, we could achieve additional benefits such as lower bitwidth per layer while recovering the accuracy completely. As a reference, Mokey~\cite{mokey} quantized the tensors of Transformer networks to 4-bit precision plus outliers without retraining. They reduced memory footprint by 50\% compared to their INT8 baseline model. In contrast, DNA-TEQ reduces the Transformer tensors' footprint by 61.86\% with an average precision of about 3 bits. Furthermore, Mokey is developed exclusively for Transformers while DNA-TEQ is an adaptive methodology that covers a wide variety of DNNs.

\subsection{Hardware Evaluation}
\begin{figure}[t!]
\centering
\includegraphics[width=0.7\columnwidth]{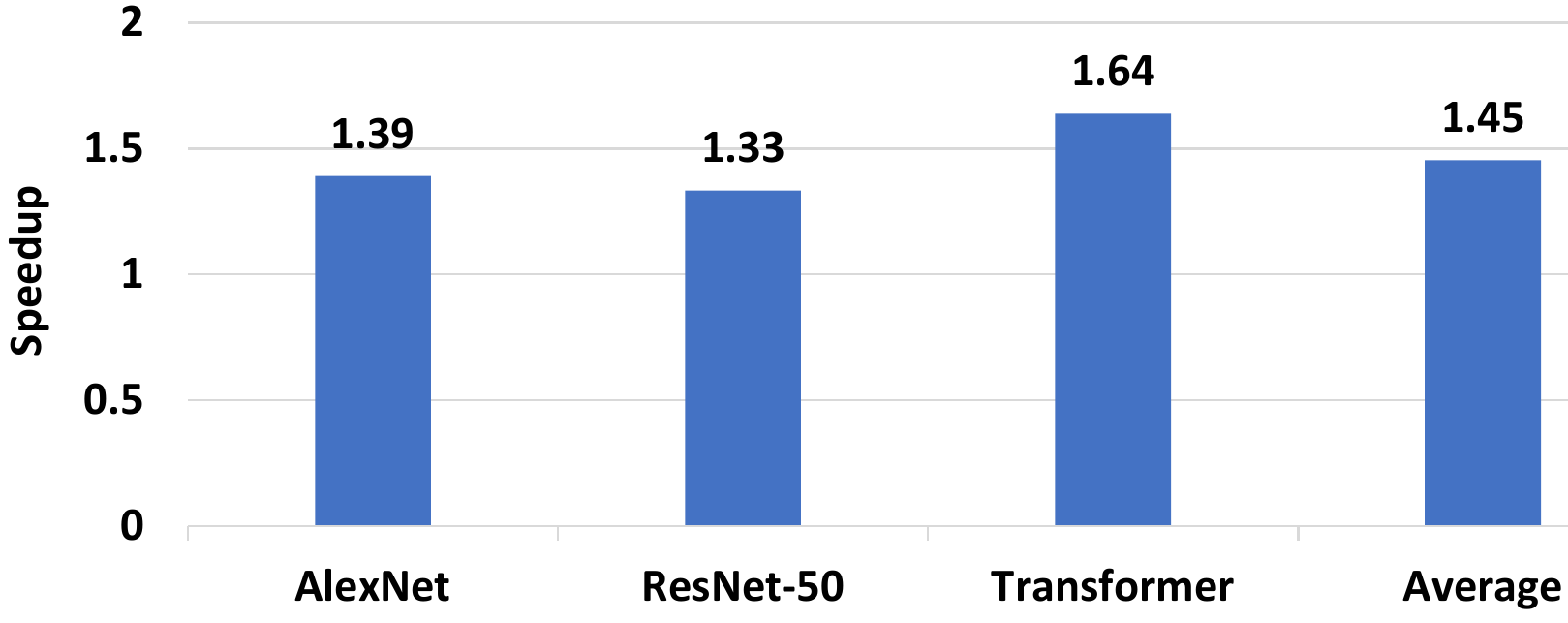}
\vskip -0.05in
\caption{Speedups of DNA-TEQ over INT8 baseline.}
\label{fig:accelerator_speedup}
\vskip -0.15in
\end{figure}

Figure~\ref{fig:accelerator_speedup} shows the speedups achieved by DNA-TEQ. Compared to the INT8 baseline accelerator, DNA-TEQ provides consistent speedups for our set of DNNs that range from $1.33x$ (\textit{ResNet}) to $1.64x$ (\textit{Transformer}), achieving an average performance improvement of $1.45x$. The reduction in execution time is due to DNA-TEQ's efficient quantization scheme. The number of memory accesses is significantly reduced since DNA-TEQ provides lower numerical precision per layer. As shown in Table~\ref{t:avg_bw_accy}, \textit{Transformer} exhibits the highest compression ratio, thus obtaining the largest performance improvement.

\begin{figure}[t!]
\centering
\includegraphics[width=0.7\columnwidth]{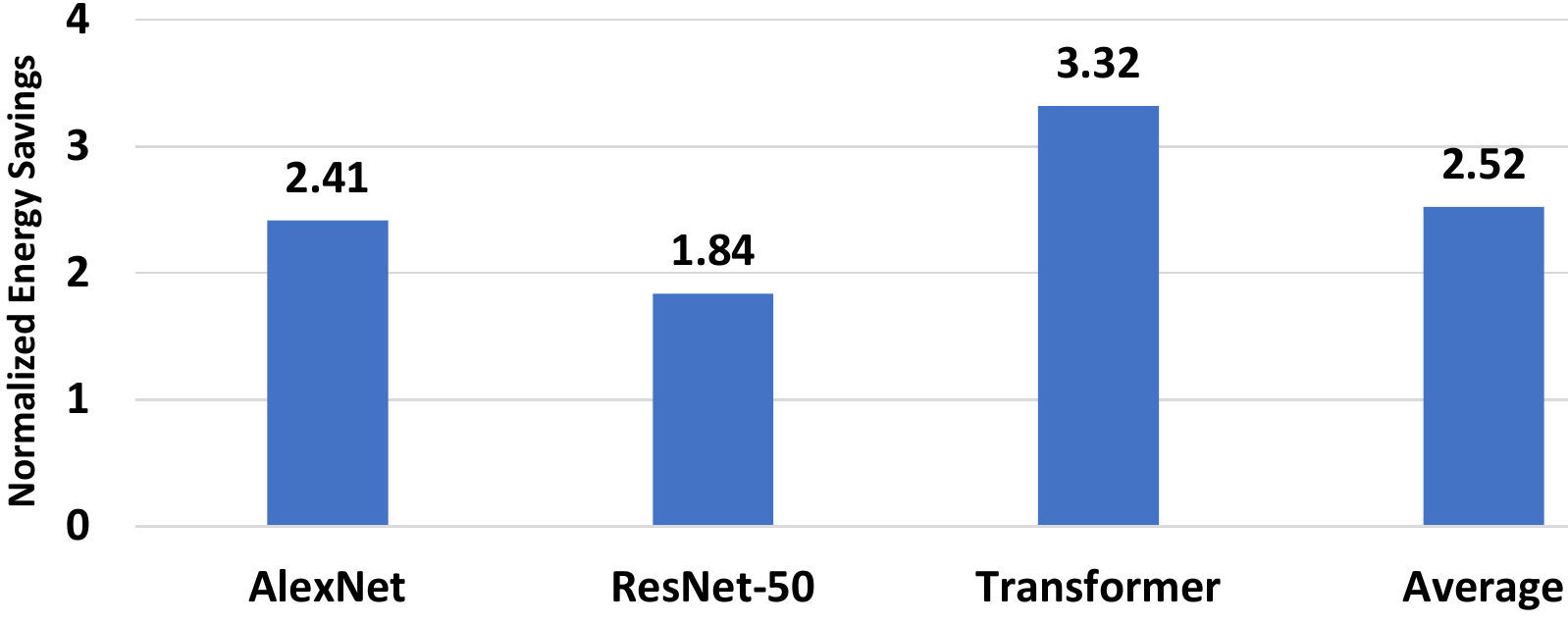}
\vskip -0.05in
\caption{Normalized energy savings for each DNN.}
\label{fig:accelerator_energy}
\vskip -0.20in
\end{figure}

Figure~\ref{fig:accelerator_energy} reports normalized energy savings. On average, DNA-TEQ reduces the energy consumption of the accelerator by $2.5x$ over the baseline. The energy savings are well correlated with the hardware simplicity of the counting of exponents with variable bitwidth per layer. These energy savings are due to two main reasons. First, dynamic energy is reduced due to the savings in multiplications and memory accesses. Second, the performance improvements shown in Figure~\ref{fig:accelerator_speedup} provide a reduction in static energy. Again, \textit{Transformer} obtains the largest benefits, achieving a reduction of $3.3x$ in energy.

\subsection{Overheads}

\begin{figure}[t!]
\centering
\includegraphics[width=0.7\columnwidth]{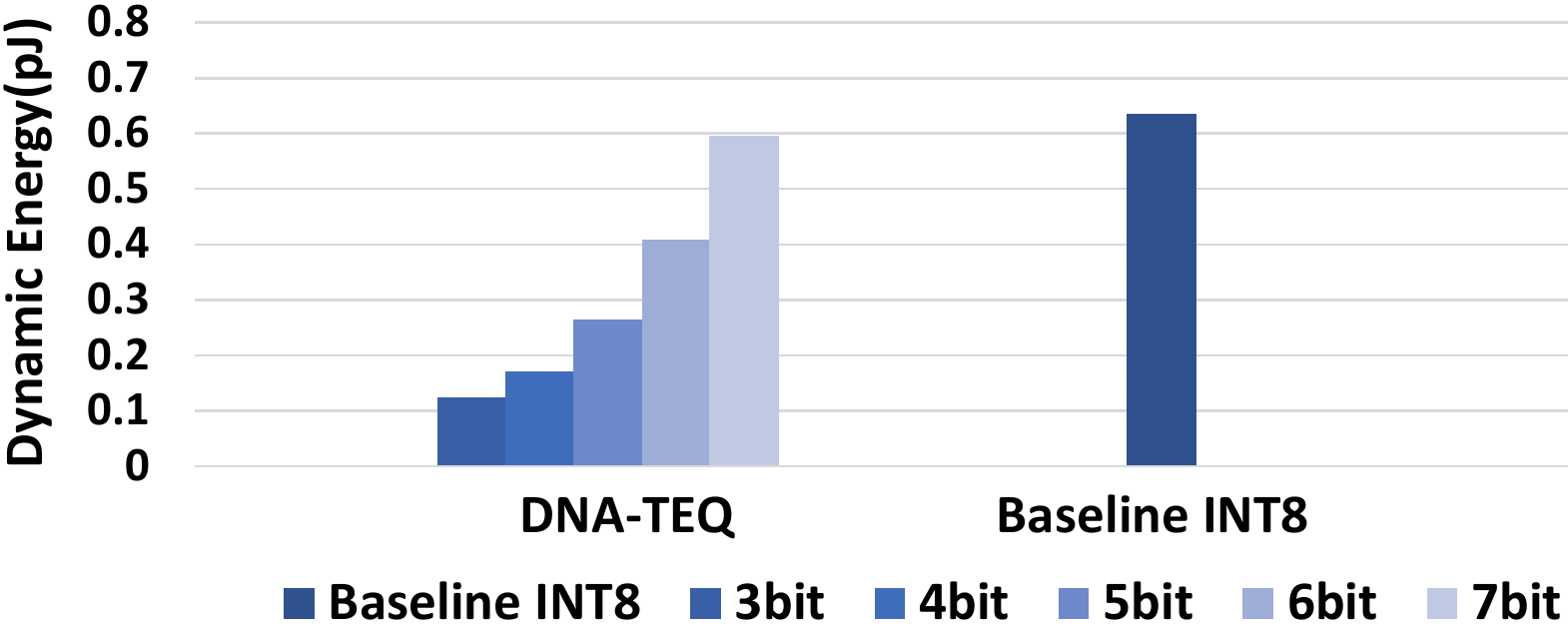}
\vskip -0.05in
\caption{Dynamic energy of a counting step with different bit-widths. The INT8 baseline accounts for the energy of a single MAC operation.}
\label{fig:energy_per_operation}
\vskip -0.20in
\end{figure}

\textbf{Energy:} Figure~\ref{fig:energy_per_operation} shows a comparison of the dynamic energy consumed by the execution of a single counting step with different quantization bitwidths. We also show the baseline equivalent with the energy consumption of an INT8 MAC operation, that is, the product and accumulation of a single input and weight. As can be seen, DNA-TEQ delivers the lowest energy consumption per operation regardless of the numerical precision. However, the main overhead of DNA-TEQ lies in the post-processing stage, which may require several FP operations according to the precision of each layer, hindering the benefits of the quantization scheme. In particular, the layers quantized with 7 bits are more energy costly than those of the INT8 baseline. These overheads are taken into account for the energy of Figure~\ref{fig:accelerator_energy}. However, the number of layers quantized with 7-bit is lower than 3\% in our set of DNNs. Therefore, the benefits due to the lower bitwidth of the remaining layers can easily overcome these modest overheads.

\textbf{Area:} DNA-TEQ replaces the MAC units by Counter-Sets to perform the exponential dot-product operations, requiring additional on-chip memory per PE. However, the extra memory does not incur in area nor energy overheads due to the MACs being more costly than the simple Counter-Sets. The area of DNA-TEQ in the logic die due to 16 PEs is $0.59mm^2$ in $32nm$. In comparison, the baseline at the same technology node occupies an area of $0.78mm^2$. In particular, the Counter-Sets of all PEs have a total area of $0.32mm^2$, while the MACs in the baseline occupy $0.67mm^2$. To ensure a fair comparison, the amount of FP multipliers required to perform de-quantization is the same in both accelerators. To summarize, despite the additional memory required by our proposal, DNA-TEQ achieves a smaller area compared to the baseline accelerator.

\subsection{Sensitivity Analysis}
The proposed quantization scheme uses two thresholds to control the bitwidth of each layer and the loss of accuracy. Thresholds $Thr_{w}$ and $Thr_{act}$ are defined as the maximum quantization error that can be introduced in the weights and activations of a given layer, respectively. Our goal is to find values for these parameters that achieve high efficiency, i.e. a large amount of compression with negligible accuracy loss, for a wide range of DNNs. Therefore, the user does not have to manually tune these thresholds for each specific DNN.

\begin{figure}[t!]
\centering
\includegraphics[width=0.9\columnwidth]{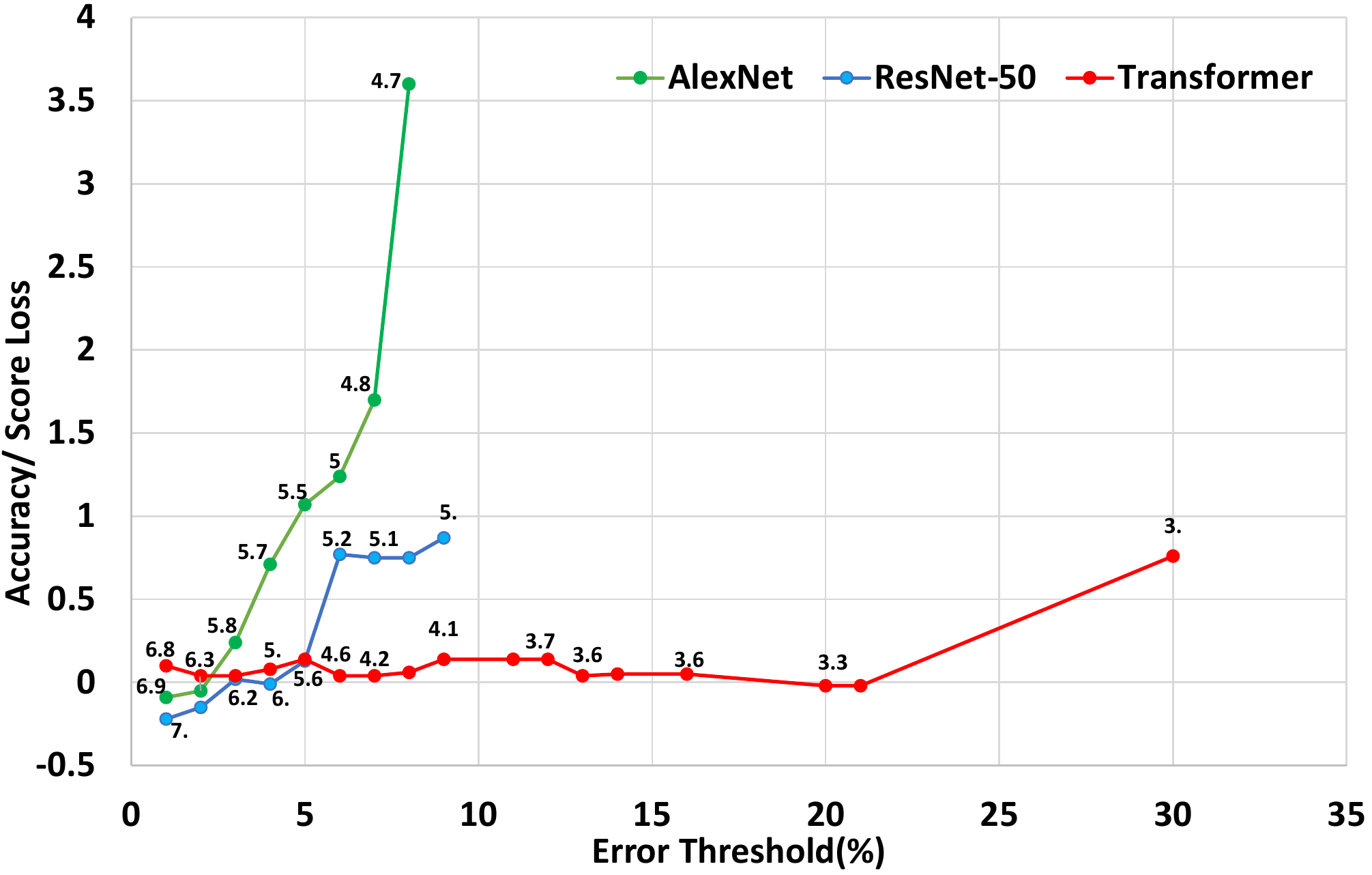}
\caption{Accuracy/Score loss respect FP32 baseline versus error threshold of Transformer, ResNet-50 and AlexNet DNNs. The number next to each point is average bitwidth obtained for each error threshold.}
\label{fig:Err_thr}
\vskip -0.2in
\end{figure}

As discussed in Section~\ref{search_alg}, $Thr_{w}$ iterates in steps of 1\% until the accuracy loss becomes higher than 1\%. In addition, we consider the first layer of each DNN as an special case that may propagate higher quantization error. Consequently, the $Thr_{w}$ of the first layer is always ten times lower than the rest. Besides, $Thr_{act}$ is computed by scaling $Thr_{w}$ to take into account the difference in magnitude between distributions, as explained above. We performed a sensitivity analysis to understand the impact of these thresholds. Figure~\ref{fig:Err_thr} shows the accuracy loss and average bitwidth of three DNNs when iterating over different error thresholds. Most of the Transformer model parameters are quantized to the lowest possible bitwidth of our scheme (i.e. 3 bits) when $Thr_{w}= 30\%$, meaning it is highly fault tolerant. The remaining parameters are quantized to either 4 or 5 bits. ResNet-50 and AlexNet reach an average bitwidth of 5.65 and 5.78 when the $Thr_{w}$ is 5\%\ and 4\%, respectively. These thresholds provide the best combination of numerical precision and DNN model accuracy. In summary, DNA-TEQ iterates over different error thresholds to achieve the best compromise between accuracy and bitwidth per layer and tensor parameters.

%% file: sections/5-conclusion.tex
\section{Conclusions}\label{conclusion}
In this paper, we propose DNA-TEQ, an approach to quantize weights and activations of DNNs in the exponential domain, which removes most of the bulky digital multipliers. This method is also motivated by the non-uniform distributions of tensors, making the exponential representation more robust and accurate compared to a linear uniform quantization. DNA-TEQ includes a search algorithm to find optimal parameters of the exponential quantization, achieving the best trade-off between numerical precision and quantization error. Results for a popular set of DNN models (Transformer, AlexNet and ResNet-50) show that, on average, DNA-TEQ provides 40\% compression with negligible accuracy loss and without retraining the DNNs. Moreover, the DNA-TEQ accelerator achieves an average performance improvement of $1.5x$ and $2.5x$ energy savings with lower area than the baseline INT8 accelerator.

%% file: sections/6-acknowledgement.tex
\section{Acknowledgement}\label{acknowledgement}
This work has been supported by the CoCoUnit ERC Advanced Grant of the EU’s Horizon 2020 program (grant No 833057), the Spanish State Research Agency (MCIN/AEI) under grant PID2020-113172RB-I00, and the ICREA Academia program.

%% file: main.bbl
\begin{thebibliography}{10}
\providecommand{\url}[1]{#1}
\csname url@samestyle\endcsname
\providecommand{\newblock}{\relax}
\providecommand{\bibinfo}[2]{#2}
\providecommand{\BIBentrySTDinterwordspacing}{\spaceskip=0pt\relax}
\providecommand{\BIBentryALTinterwordstretchfactor}{4}
\providecommand{\BIBentryALTinterwordspacing}{\spaceskip=\fontdimen2\font plus
\BIBentryALTinterwordstretchfactor\fontdimen3\font minus
  \fontdimen4\font\relax}
\providecommand{\BIBforeignlanguage}[2]{{%
\expandafter\ifx\csname l@#1\endcsname\relax
\typeout{** WARNING: IEEEtranS.bst: No hyphenation pattern has been}%
\typeout{** loaded for the language `#1'. Using the pattern for}%
\typeout{** the default language instead.}%
\else
\language=\csname l@#1\endcsname
\fi
#2}}
\providecommand{\BIBdecl}{\relax}
\BIBdecl

\bibitem{language}
T.~Brown, B.~Mann, N.~Ryder, M.~Subbiah, J.~D. Kaplan, P.~Dhariwal,
  A.~Neelakantan, P.~Shyam, G.~Sastry, A.~Askell \emph{et~al.}, ``Language
  models are few-shot learners,'' \emph{Advances in neural information
  processing systems (NIPS)}, vol.~33, pp. 1877--1901, 2020.

\bibitem{deep}
J.~Cai, M.~Takemoto, and H.~Nakajo, ``A deep look into logarithmic quantization
  of model parameters in neural networks,'' in \emph{Proceedings of the 10th
  International Conference on Advances in Information Technology (IAIT)}, 2018,
  pp. 1--8.

\bibitem{survey}
Y.~Cheng, D.~Wang, P.~Zhou, and T.~Zhang, ``A survey of model compression and
  acceleration for deep neural networks,'' \emph{arXiv}, 2017.

\bibitem{2022log_asap}
M.~Christ, F.~de~Dinechin, and F.~P{\'e}trot, ``Low-precision logarithmic
  arithmetic for neural network accelerators,'' in \emph{33rd IEEE
  International Conference on Application-specific Systems, Architectures and
  Processors (ASAP)}, 2022.

\bibitem{permdnn}
C.~Deng, S.~Liao, Y.~Xie, K.~K. Parhi, X.~Qian, and B.~Yuan, ``Permdnn:
  Efficient compressed dnn architecture with permuted diagonal matrices,'' in
  \emph{51st Annual IEEE/ACM international symposium on microarchitecture
  (MICRO)}, 2018, pp. 189--202.

\bibitem{switch_transformer}
W.~Fedus, B.~Zoph, and N.~Shazeer, ``Switch transformers: Scaling to trillion
  parameter models with simple and efficient sparsity,'' 2021.

\bibitem{tetris}
M.~Gao, J.~Pu, X.~Yang, M.~Horowitz, and C.~Kozyrakis, ``Tetris: Scalable and
  efficient neural network acceleration with 3d memory,'' in \emph{Proceedings
  of the Twenty-Second International Conference on ASPLOS}, 2017, pp. 751--764.

\bibitem{eie}
S.~Han, X.~Liu, H.~Mao, J.~Pu, A.~Pedram, M.~A. Horowitz, and W.~J. Dally,
  ``Eie: efficient inference engine on compressed deep neural network,'' in
  \emph{Proceedings of the 43rd ISCA}, 2016, pp. 243--254.

\bibitem{ResNet}
K.~He, X.~Zhang, S.~Ren, and J.~Sun, ``Deep residual learning for image
  recognition,'' in \emph{IEEE Conference on Computer Vision and Pattern
  Recognition (CVPR)}, 2016, pp. 770--778.

\bibitem{VNNI}
Intel, ``Deep learning with intel® avx-512 \& intel® dl boost,''
  \url{https://www.intel.com/content/www/us/en/developer/articles/guide/deep-learning-with-avx512-and-dl-boost.html}.

\bibitem{khoram-adaptive}
S.~Khoram and J.~Li, ``Adaptive quantization of neural networks,'' in
  \emph{International Conference on Learning Representations (ICLR)}, 2018.

\bibitem{neurocube}
D.~Kim, J.~Kung, S.~Chai, S.~Yalamanchili, and S.~Mukhopadhyay, ``Neurocube: A
  programmable digital neuromorphic architecture with high-density 3d memory,''
  in \emph{Proceedings of the 43rd International Symposium on Computer
  Architecture (ISCA)}, 2016, p. 380–392.

\bibitem{quantizing}
R.~Krishnamoorthi, ``Quantizing deep convolutional networks for efficient
  inference: A whitepaper,'' \emph{arXiv}, 2018.

\bibitem{AlexNetV2}
A.~Krizhevsky, ``One weird trick for parallelizing convolutional neural
  networks,'' \emph{arXiv}, 2014.

\bibitem{understanding}
G.~Li, S.~K.~S. Hari, M.~Sullivan, T.~Tsai, K.~Pattabiraman, J.~Emer, and S.~W.
  Keckler, ``Understanding error propagation in deep learning neural network
  (dnn) accelerators and applications,'' in \emph{Proceedings of the
  International Conference for High Performance Computing, Networking, Storage
  and Analysis (SC)}, 2017, pp. 1--12.

\bibitem{additive}
Y.~Li, X.~Dong, and W.~Wang, ``Additive powers-of-two quantization: An
  efficient non-uniform discretization for neural networks,'' in
  \emph{International Conference on Learning Representations (ICLR)}, 2020.

\bibitem{recurrent}
T.~Mikolov, M.~Karafi{\'a}t, L.~Burget, J.~Cernock{\`y}, and S.~Khudanpur,
  ``Recurrent neural network based language model.'' in \emph{Interspeech},
  vol.~2, no.~3.\hskip 1em plus 0.5em minus 0.4em\relax Makuhari, 2010, pp.
  1045--1048.

\bibitem{convolutional}
D.~Miyashita, E.~H. Lee, and B.~Murmann, ``Convolutional neural networks using
  logarithmic data representation,'' \emph{arXiv}, 2016.

\bibitem{datafree}
M.~Nagel, M.~v. Baalen, T.~Blankevoort, and M.~Welling, ``Data-free
  quantization through weight equalization and bias correction,'' in
  \emph{Proceedings of the IEEE/CVF International Conference on Computer Vision
  (ICCV)}, 2019, pp. 1325--1334.

\bibitem{bit_eff_qt}
P.~Nayak, D.~Zhang, and S.~Chai, ``Bit efficient quantization for deep neural
  networks,'' in \emph{Workshop on Energy Efficient Machine Learning and
  Cognitive Computing-NeurIPS Edition (EMC2-NIPS)}, 2019, pp. 52--56.

\bibitem{algorithm_architecture}
A.~Pedram, A.~S. Ardestani, L.~Li, H.~Abdelaziz, J.~Fang, and J.~Hassoun,
  ``Algorithm/architecture solutions to improve beyond uniform quantization in
  embedded dnn accelerators,'' \emph{Journal of Systems Architecture (JSA)},
  vol. 126, p. 102454, 2022.

\bibitem{mriera_thesis}
M.~Riera, ``Low-power accelerators for cognitive computing,'' 2020.

\bibitem{marc_prune}
M.~Riera, J.~M. Arnau, and A.~Gonz{\'a}lez, ``Dnn pruning with principal
  component analysis and connection importance estimation,'' \emph{Journal of
  Systems Architecture (JSA)}, vol. 122, p. 102336, 2022.

\bibitem{rokh2022comprehensive}
B.~Rokh, A.~Azarpeyvand, and A.~Khanteymoori, ``A comprehensive survey on model
  quantization for deep neural networks,'' \emph{arXiv}, 2022.

\bibitem{redy}
M.~Sabri, M.~Riera, and A.~Gonz{\'a}lez, ``Redy: A novel reram-centric dynamic
  quantization approach for energy-efficient cnn inference,'' \emph{arXiv
  preprint arXiv:2306.16298}, 2023.

\bibitem{song}
Z.~Song, B.~Fu, F.~Wu, Z.~Jiang, L.~Jiang, N.~Jing, and X.~Liang, ``Drq:
  Dynamic region-based quantization for deep neural network acceleration,'' in
  \emph{ACM/IEEE 47th Annual ISCA}, 2020, pp. 1010--1021.

\bibitem{deeptest}
Y.~Tian, K.~Pei, S.~Jana, and B.~Ray, ``Deeptest: Automated testing of
  deep-neural-network-driven autonomous cars,'' in \emph{International
  conference on software engineering (ICSE)}, 2018, pp. 303--314.

\bibitem{L2L}
S.~Ullah, S.~Gupta, K.~Ahuja, A.~Tiwari, and A.~Kumar, ``L2l: A highly accurate
  log\_2\_lead quantization of pre-trained neural networks,'' in \emph{IEEE
  DATE}, 2020, pp. 979--982.

\bibitem{attention}
A.~Vaswani, N.~Shazeer, N.~Parmar, J.~Uszkoreit, L.~Jones, A.~N. Gomez,
  {\L}.~Kaiser, and I.~Polosukhin, ``Attention is all you need,''
  \emph{Advances in neural information processing systems (NeurIPS)}, vol.~30,
  2017.

\bibitem{pre-trained}
S.~Vogel, J.~Springer, A.~Guntoro, and G.~Ascheid, ``Self-supervised
  quantization of pre-trained neural networks for multiplierless
  acceleration,'' in \emph{IEEE DATE}, 2019, pp. 1094--1099.

\bibitem{google_human}
Y.~Wu, M.~Schuster, Z.~Chen, Q.~V. Le, M.~Norouzi, W.~Macherey, M.~Krikun,
  Y.~Cao, Q.~Gao, K.~Macherey \emph{et~al.}, ``Google's neural machine
  translation system: Bridging the gap between human and machine translation,''
  \emph{arXiv}, 2016.

\bibitem{mokey}
A.~H. Zadeh, M.~Mahmoud, A.~Abdelhadi, and A.~Moshovos, ``Mokey: Enabling
  narrow fixed-point inference for out-of-the-box floating-point transformer
  models,'' in \emph{IEEE Proceedings of the 49th Annual ISCA}, 2022, p.
  888–901.

\bibitem{transpim}
M.~Zhou, W.~Xu, J.~Kang, and T.~Rosing, ``Transpim: A memory-based acceleration
  via software-hardware co-design for transformer,'' in \emph{IEEE
  International Symposium on HPCA}, 2022, pp. 1071--1085.

\end{thebibliography}
